%% file: main.tex
\documentclass[lettersize,journal]{IEEEtran}
\usepackage{amsmath,amsfonts}
\usepackage{algorithmic}
\usepackage{algorithm}
\usepackage{array}
\usepackage{textcomp}
\usepackage{stfloats}
\usepackage{url}
\usepackage{verbatim}
\usepackage{graphicx}
\usepackage{hyperref}
\usepackage{amsmath,amssymb}
\usepackage[dvipsnames]{xcolor}
\usepackage{graphicx} 
\usepackage{pgf} 
\usepackage{pgfplots}
\usepackage{booktabs}
\usepackage[caption=false,font=footnotesize]{subfig}
\usepackage{multirow}
\usepackage{float}

\usepackage{pifont} 

\usepackage[utf8x]{inputenc}
\usepackage{tikz}
\usepackage{tikz-cd}
\usetikzlibrary{snakes,arrows,shapes,positioning,arrows.meta}

\usepackage[most]{tcolorbox}
\tcbuselibrary{listings, skins}

\usetikzlibrary{arrows.meta}

\tikzset{
    vertex/.style = {circle, draw, minimum size=1.5em, inner sep=1pt},
    edge/.style = {line width=0.5mm},
    arrow/.style = {-{Latex[length=3mm, width=2mm]}, edge},
    biarrow/.style = {Latex-Latex, edge},
    oarrow/.style = {o-{Latex[length=3mm, width=2mm]}, edge},
    oend/.style = {o-o, edge},
    larrow/.style = {-{Latex[length=3mm, width=2mm]}, edge},
    barrow/.style = {{Latex[length=3mm, width=2mm]}-, edge}
}

\usepackage{svg}
\usepackage{balance}

\newcommand{\indep}{\perp\!\!\!\perp}
\newcommand{\nindep}{\not\!\perp\!\!\!\perp}
\def\*#1{\mathbf{#1}}

\tikzset{every picture/.style={scale=1.5}}

\usepackage[
backend=biber,
style=ieee,
sorting=ynt
]{biblatex}
\AtEveryBibitem{%
  \clearfield{issn}%
  \clearfield{isbn}%
  \clearfield{abstract}%
  \clearfield{file}%
  \clearfield{eprint}%
  \clearlist{language}%
  \clearfield{note}%
  \ifentrytype{article}{
    \clearfield{url}%
    \clearfield{urldate}%
  }{}%
  \clearfield{month}
  \ifentrytype{book}{}{
    \clearlist{location}
  }%
  \ifentrytype{article}{\clearfield{url}}{}
  \ifentrytype{inproceedings}{\clearfield{url}}{}
  \iffieldundef{doi}
    {} 
    {\clearfield{url}} 
}

\AtBeginBibliography{\small}

\begin{document}

\title{Federated Causal Discovery Across Heterogeneous Datasets under Latent Confounding}

\author{Maximilian Hahn$^{1}$, Alina Zajak$^{1}$, Dominik Heider$^{1}$, and Ad\`ele H. Ribeiro$^{1}$
	\thanks{
		$^{1}$University of Münster, Institute of Medical Informatics, Münster, Germany
	}
}

\markboth{Journal of \LaTeX\ Class Files,~Vol.~14, No.~8, August~2021}%
{Hahn \MakeLowercase{\textit{et al.}}: Federated Causal Discovery Across Heterogeneous Datasets under Latent Confounding}


\maketitle

\begin{abstract}
	Causal discovery across multiple datasets is often constrained by data privacy regulations
	and cross-site heterogeneity, limiting the use of conventional methods that require
	a single, centralized dataset. To address these challenges, we introduce
	\emph{fedCI}, a federated conditional independence test
	that rigorously handles heterogeneous datasets with
	non-identical sets of variables, site-specific effects, and
	mixed variable types, including continuous, ordinal, binary, and categorical variables.
	At its core, fedCI uses a federated Iteratively Reweighted Least Squares (IRLS)
	procedure to estimate the parameters of generalized linear models underlying
	likelihood-ratio tests for conditional independence.
	Building on this, we develop \emph{fedCI-IOD}, a federated extension of the
	Integration of Overlapping Datasets (IOD) algorithm,
	that replaces its meta-analysis strategy and enables, for the fist time,
	federated causal discovery under latent confounding across distributed and heterogeneous datasets.
	By aggregating evidence federatively, fedCI-IOD
	not only preserves privacy but also substantially enhances statistical power,
	achieving performance comparable to fully pooled analyses and
	mitigating artifacts from low local sample sizes.
	Our tools are publicly available as the \emph{fedCI} Python package,
	a privacy-preserving R implementation of IOD,
	and a web application for the \emph{fedCI-IOD} pipeline,
	providing versatile, user-friendly solutions for federated
	conditional independence testing and causal discovery.\looseness=-1
	%
\end{abstract}


\input{01introduction.tex}
\input{02background.tex}
\input{03methodology.tex}
\input{04simulations.tex}
\input{05conclusion.tex}


\section{Acknowledgements}
AHR acknowledges support from the
Federal Ministry of Research, Technology and Space in Germany (BMFTR) through the grant
CausalAI4Health 
(reference 01ZU2503).

\nocite{*}
\printbibliography

\vspace{-0.5em}
\section{Biography Section}
\vspace{-3em}
\begin{IEEEbiographynophoto}{Maximilian Hahn}
	is a Ph.D. student in the CausalAI4Health research group at the Institute of Medical Informatics, University of Münster.
	He received the M.Sc. degree in Data Science from the University of Marburg (2023) and the B.Sc. degree in Computer Science from DHBW Horb (2020).
	His research focuses on Machine Learning and causal discovery.
\end{IEEEbiographynophoto}
\vspace{-3em}
\begin{IEEEbiographynophoto}{Alina Zajak}
	has been working as a product technician at HDI Insurance since October 2024.
	Prior to this, she completed Bachelor's degrees in Mathematics and Computer Science at Heinrich Heine University Düsseldorf (2019–2024).
\end{IEEEbiographynophoto}
\vspace{-3em}
\begin{IEEEbiographynophoto}{Dominik Heider}
	has been director of the Institute of Medical Informatics,
	University of Münster, since 2024.
	He was a professor at University of Düsseldorf (2023-2024), Marburg (2016-2023),
	and TUM Campus Straubing (2014-2016), and a visiting professor at Harvard University
	since 2023. He holds a diploma (2006) and PhD (2008) in computer science from
	Münster and a habilitation (2012) from Duisburg-Essen.
	His research focuses on developing AI and Machine Learning tools for biomedical challenges.
\end{IEEEbiographynophoto}
\vspace{-3em}
\begin{IEEEbiographynophoto}{Adèle Ribeiro}
	is head of the CausalAI4Health research group at the Institute of Medical Informatics,
	University of Münster. Previously, she worked at the University of Marburg (2022-2024),
	Columbia University (2019-2022), and the Heart Institute at
	University of São Paulo (USP). She holds a Ph.D. and M.Sc. in Computer Science and
	a B.Sc. in Mathematics from USP.
	Her work focuses on advancing AI and Machine Learning through
	causal reasoning.\looseness=-1
\end{IEEEbiographynophoto}


\begin{appendices}
	\input{99appendix_A.tex}

	\input{99appendix_B.tex}
	\input{99appendix_C.tex}
	\input{99appendix_D.tex}
	\input{99appendix_E.tex}
	\input{99appendix_F.tex}
\end{appendices}

\end{document}

%% file: 01introduction.tex
\section{Introduction}

\IEEEPARstart{I}{n} numerous scientific and industrial domains, from healthcare to economics,
understanding the relationships between variables is paramount.
A key aspect of this is discerning
conditional
independence (CI) between variables, which
plays a central role in tasks such as
feature selection in machine learning or for edge orientation in causal discovery.
To assess whether two (sets of) variables are conditionally independent given others,
researchers typically employ CI tests.
The reliability of these tests depends not only on controlling type I error, set by the significance level,
but also on minimizing type II errors through adequate statistical power,
determined by sample size and effect strength.
Importantly, failing to reject the null hypothesis of independence does not
confirm independence and may merely reflect insufficient sensitivity.
Thus, while dependencies can often be identified with confidence,
independencies can be inferred reliably only when the test is
sufficiently powerful to rule out even the weakest alternatives.

CI testing is fundamental to causal discovery.
In particular, constraint-based causal discovery
algorithms infer causal relationships among observed variables by
systematically evaluating conditional independencies across relevant variable sets,
with each test guiding the presence, absence, or orientation of edges within the underlying causal graph.
Their correctness, however, relies on the \emph{faithfulness} assumption,
which requires that all observed conditional independencies correspond exactly to
those implied by the underlying causal graph.
In this setting, misinterpreting non-rejection of independence as evidence of independence
is critical.
A single false inference can propagate through the algorithm,
triggering a cascade of errors and yielding an entirely incorrect causal structure.

Despite the critical need for large datasets to ensure sufficient statistical power in CI tests,
acquiring and centralizing such data poses considerable challenges.
In many fields, such as medicine, generating new data is often difficult or costly,
necessitating reliance on existing datasets.
However, combining multi-center data sources into a single dataset
is often hindered by various challenges, including
incomplete overlap of observed variables across sites and
strict legal and ethical constraints (such as data protection regulations), especially when handling sensitive information.

On the one hand, methods such as meta-analysis can synthesize findings from separate studies, but
they operate on summary statistics, not fully exploiting
the information in the data.
Federated learning, on the other hand, offers a compelling alternative, enabling statistical analysis
across distributed datasets while preserving privacy~\cite{Yang2019, Hauschild2022, Tajabadi2024}.

Although federated causal discovery has seen notable progress, existing frameworks remain limited by restrictive assumptions
on 
data structure and causal sufficiency.
Most current approaches, such as FedDAG \cite{Gao2023} and Fed$C^2$SL~\cite{Wang2023, Wang2021},
rely on the premise of causal sufficiency, which assumes that all confounding variables are observed.
Furthermore, these methods often require identical variable sets across all sites, with Fed$C^2$SL being further limited to categorical data.
While FedCDH~\cite{Li2024} accommodates mixed-type data, it still necessitates fully overlapping variable sets.
This requirement is particularly restrictive in real-world multi-center analyses where data is often both horizontally and vertically partitioned.
While FedISHC~\cite{Chen2025} and FedPuzzle~\cite{Li2025} attempt to relax these overlap constraints, they continue to assume causal sufficiency,
which limits their applicability in real-world environments.

Beyond these theoretical constraints, a significant implementation gap persists in the literature.
To our knowledge, existing federated causal discovery methods lack publicly available software implementations that include essential features for real-world deployment,
such as robust network communication protocols.
Consequently, no current method can simultaneously accommodate non-identical sets of variables, handle mixed-type data,
model site-specific effects, account for latent confounding, and provide a deployable software framework.
This leaves a substantial methodological and practical gap for distributed data analysis, which we formalize as follows:\looseness=-1

\vspace{0.4em}
\noindent\textbf{Problem Statement.}
Given distributed, observational datasets with mixed variable types and subject to strict privacy constraints,
site-specific effects, non-identical variable sets, and latent confounding,
the goal is to reliably infer the underlying causal relationships.
A secondary objective is to provide a robust, computationally efficient software implementation to ensure the practical viability of the proposed method.
\vspace{0.4em}

To address this problem, we introduce \emph{fedCI}, the first federated CI testing framework
tailored to distributed datasets with heterogeniety in the form of
non-identical variable sets, mixed data types, and site-specific effects,
supported by a fully operational Python implementation.
Building on this, we provide a federated extension of the Integration of Overlapping Datasets (IOD)
algorithm~\cite{Tillman2011}. Under the sole assumption of faithfulness, IOD is
provably sound and complete for causal discovery from multiple datasets with
non-identical variable sets in the presence of latent confounding.
It infers a list of Partial Ancestral Graphs (PAGs), each representing a
Markov equivalence class (MEC) of causal models consistent with the conditional
independencies observed across all datasets.
In a PAG, a tail at an edge indicates the variable is an ancestor (i.e., has a causal path)
to the other variable in every model of the MEC, an arrowhead indicates it is never an ancestor, and
a circle ($\circ$) denotes a relationship that varies within the MEC.

In its original form, IOD combines CI test results across datasets via  Fisher's method~\cite{Fisher1925}.
To our knowledge, it has only been
implemented in the TETRAD software suite~\cite{ramsey2018tetrad}, 
where it requires centralized access to all datasets
and therefore does not protect sensitive local data.

We provide the first privacy-preserving implementation of IOD as an R package,
modifying the algorithm to operate safely across multiple datasets
either via meta-analysis based solely on shared test statistics or $p$-values,
or through seamless integration with federated CI tests.
Building on this, and
leveraging fedCI, our \emph{fedCI-IOD} framework is the first
to enable federated causal discovery across heterogeneous datasets
-- with non-identical variable sets, site-specific effects, and mixed variable types --
under latent confounding, retaining
IOD's theoretical guarantees while rigorously preserving data privacy and
fully utilizing available information.
These advances substantially increase statistical power, offering a practical
and rigorous solution for learning causal structures from distributed datasets
without any need for data pooling.

To construct the fedCI framework,
we employ Likelihood-Ratio Tests (LRTs) for their strong theoretical guarantees
and minimal assumptions, framing CI assessment as a comparison between two
appropriately specified nested models. 
As the modeling basis for our LRTs, we adopt Generalized Linear Models (GLMs),
which offer an exceptional combination of interpretability and statistical robustness,
with demonstrated applicability across diverse real-world tasks.
Although linear in their parameters, GLMs can capture complex,
nonlinear relationships through suitable transformations or basis
expansions of the predictors.
They also support a wide range of outcome types
through appropriate model specifications, including
Gaussian models for continuous outcomes,
(multinomial) logistic models for categorical outcomes, and
(generalized) ordered logit models for ordinal outcomes.

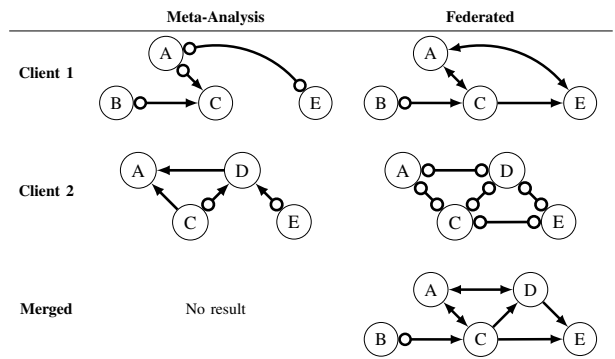
\begin{figure}
	\centering
	\scalebox{.6}{
		\begin{tabular}{c c c}
			                  & \textbf{Meta-Analysis}         & \textbf{Federated} \\
			\midrule
			\textbf{Client 1} & \resizebox{0.3\textwidth}{!}{
				\begin{tikzpicture}[node distance=0.75cm, auto, baseline=(current bounding box.center)]
					\tikzset{vertex/.style = {shape=circle, draw, minimum size=1.5em}}
					\node (B) [circle, draw] {B};
					\node (A) [circle, draw, above right=0.7cm of B] {A};
					\node (C) [circle, draw, below right=0.7cm of A] {C};
					\node (E) [circle, draw, right=1.3cm of C] {E};

					\draw[o-latex, line width=0.5mm] (A) -- (C);
					\draw[o-o, line width=0.5mm, bend left=30] (A) to (E);
					\draw[o-latex, line width=0.5mm] (B) -- (C);
				\end{tikzpicture}
			}                 & \resizebox{0.3\textwidth}{!}{
				\begin{tikzpicture}[node distance=0.75cm, auto, baseline=(current bounding box.center)]
					\tikzset{vertex/.style = {shape=circle, draw, minimum size=1.5em}}
					\node (B) [circle, draw] {B};
					\node (A) [circle, draw, above right=0.7cm of B] {A};
					\node (C) [circle, draw, below right=0.7cm of A] {C};
					\node (E) [circle, draw, right=1.3cm of C] {E};

					\draw[latex-latex, line width=0.5mm] (A) -- (C);
					\draw[latex-latex, line width=0.5mm, bend left=30] (A) to (E);
					\draw[o-latex, line width=0.5mm] (B) -- (C);
					\draw[-latex, line width=0.5mm] (C) -- (E);
				\end{tikzpicture}
			}
			\\
			\addlinespace[2em]
			\textbf{Client 2} & \resizebox{0.25\textwidth}{!}{
				\begin{tikzpicture}[node distance=0.75cm, auto, baseline=(current bounding box.center)]
					\tikzset{vertex/.style = {shape=circle, draw, minimum size=1.5em}}
					\node (A) [circle, draw] {A};
					\node (C) [circle, draw, below right=0.7cm of A] {C};
					\node (D) [circle, draw, above right=0.7cm of C] {D};
					\node (E) [circle, draw, below right=0.7cm of D] {E};

					\draw[latex-, line width=0.5mm] (A) -- (C);
					\draw[latex-, line width=0.5mm] (A) -- (D);
					\draw[o-latex, line width=0.5mm] (C) -- (D);
					\draw[latex-o, line width=0.5mm] (D) -- (E);
				\end{tikzpicture}
			}                 & \resizebox{0.25\textwidth}{!}{
				\begin{tikzpicture}[node distance=0.75cm, auto, baseline=(current bounding box.center)]
					\tikzset{vertex/.style = {shape=circle, draw, minimum size=1.5em}}
					\node (A) [circle, draw] {A};
					\node (C) [circle, draw, below right=0.7cm of A] {C};
					\node (D) [circle, draw, above right=0.7cm of C] {D};
					\node (E) [circle, draw, below right=0.7cm of D] {E};

					\draw[o-o, line width=0.5mm] (A) -- (C);
					\draw[o-o, line width=0.5mm] (A) -- (D);
					\draw[o-o, line width=0.5mm] (C) -- (D);
					\draw[o-o, line width=0.5mm] (C) -- (E);
					\draw[o-o, line width=0.5mm] (D) -- (E);
				\end{tikzpicture}
			}
			\\
			\addlinespace[2em]
			\textbf{Merged}   &
			No result         & \resizebox{0.3\textwidth}{!}{
				\begin{tikzpicture}[node distance=0.75cm, auto, baseline=(current bounding box.center)]
					\tikzset{vertex/.style = {shape=circle, draw, minimum size=1.em}}
					\node (B) [circle, draw] {B};
					\node (A) [circle, draw, above right=0.7cm of B] {A};
					\node (C) [circle, draw, below right=0.7cm of A] {C};
					\node (D) [circle, draw, above right=0.7cm of C] {D};
					\node (E) [circle, draw, below right=0.7cm of D] {E};

					\draw[o-latex, line width=0.5mm] (B) -- (C);
					\draw[-latex, line width=0.5mm] (C) -- (E);
					\draw[-latex, line width=0.5mm] (C) -- (D);
					\draw[latex-latex, line width=0.5mm] (A) -- (C);
					\draw[latex-latex, line width=0.5mm] (A) -- (D);
					\draw[latex-, line width=0.5mm] (E) -- (D);
				\end{tikzpicture}
			}
		\end{tabular}
	}
	\caption{
		Comparison of constraint-based causal discovery results of the IOD algorithm in its original form using meta-analysis~\cite{Tillman2011}
		and our proposed federated adaptation.
		Differences in the CI tests led to different local graphs for each client.
		Merging these local results into one PAG that adheres to all constraints was impossible for meta-analysis,
		but our federated approach was able to find the correct data-generating PAG.
	}
	\label{fig:intro example}
\end{figure}

\autoref{fig:intro example} highlights the critical need to
boost statistical power in causal discovery through federated CI tests,
while simultaneously accommodating datasets that differ in their observed
variables and may be affected by latent confounding.
In this example, the original IOD algorithm
fails to produce a PAG over the full set of all observed variables.
This occurs because CI tests on the local datasets lack sufficient statistical power,
leading to incorrect CI decisions and misoriented local PAGs that would not arise with pooled data.
In contrast, fedCI-IOD, which integrates the adapted IOD algorithm with fedCI,
successfully captures global statistical evidence across all clients, yielding
accurate CI decisions and enabling recovery of the true underlying local and
full PAGs without compromising data privacy.
Notably, this example also underscores the importance of accommodating
non-identical sets of variables: several definitive relationships
(e.g., $C \rightarrow D$, $D \rightarrow E$, and $A \leftrightarrow D$)
could only be determined by inferring constraints among variables that were
not observed within the same dataset, such as establishing that
$B \nindep D$ but $B \indep D \mid C$.

\begin{figure*}
	\centering
	\includesvg[width=0.8\linewidth,inkscapelatex=true,pretex=\scriptsize]{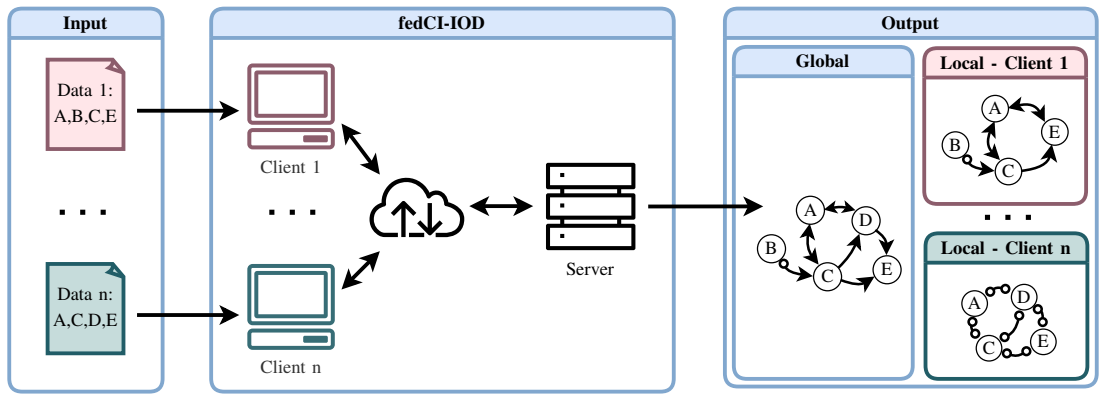}
	\caption{
		Schematic overview of federated causal discovery, including clients with non-identical variable sets
		and a managing server which are in potentially seperate networks, obtaining individual as well as unified PAGs.
		In Appendix~\ref{sec:webapp appx}, screenshots of our web application showcase each step of this process.
	}
	\label{fig:fedci iod schema}
\end{figure*}

Until now, there has been no readily available application that enables
multiple parties to collaboratively perform federated causal discovery or
contribute local data to a global causal model.
Our fedCI-IOD framework is delivered through a portable client--server web application,
making federated causal discovery accessible for the first time.
This platform allows site-specific data to be combined into a unified
causal model across all contributors, while rigorously preserving the privacy of local datasets.
A schematic overview of our approach is provided in \autoref{fig:fedci iod schema}, while 
real screenshots of our application are provided in Appendix~\ref{sec:webapp appx}, \autoref{fig:webapp 1}.

This work advances the state of the art in federated CI testing and causal discovery through three key contributions:
\begin{enumerate}
	\item We introduce \emph{fedCI}, the first federated CI testing framework specifically designed
	      for distributed datasets with non-identical variable sets, mixed data types, and site-specific effects.
	      It is made available as a Python package which provides a privacy-preserving client--server architecture with network-communication
	      and efficient computations across multiple data holders.

	\item We provide a privacy-preserving implementation of the IOD algorithm as
	      an R package for safe, collaborative causal discovery, operating either via meta-analysis
	      of shared $p$-values or seamless integration with federated CI tests,
	      while retaining all theoretical guarantees. \looseness=-1

	\item We develop a fully containerized client--server web platform for the \emph{fedCI–IOD} pipeline
	      that enables, for the first time, federated causal discovery across heterogeneous datasets featuring
	      non-identical sets of variables, mixed data types, site-specific effects, and latent confounding, within a privacy-preserving and easily
	      deployable environment for real-world analyses.
\end{enumerate}

%% file: 02background.tex
\section{Background}
\label{sec:background}

We first introduce the notation.
Random variables are denoted by capital letters (e.g., $X$), and their realizations by lowercase letters (e.g., $X = x$).
Boldface letters denote sets of random variables (e.g., $\*{X}$).
For random variables $X$ and $Y$ and a (possibly empty) conditioning set $\*{Z}$,
we write $X \indep Y \mid \*{Z}$ to denote conditional independence and $X \nindep Y \mid \*{Z}$ otherwise.
If $\*{Z} = \emptyset$, the conditioning set is omitted and we simply write $X \indep Y$ or $X \nindep Y$.
The following subsections present fundamental concepts used throughout the paper. \looseness=-1

\subsection{LRTs for Assessing Conditional Independence}
\label{sec:background lrt}
LRTs provide a rigorous and flexible framework for assessing conditional independence
by comparing the goodness of fit of two nested statistical models~\cite{Glover2004}.
To test a CI statement such as $X \indep Y \mid \*{Z}$, one specifies two nested models:
a restricted (null) model, $M_0$, that encodes the independence constraint (e.g., a model for $Y$ using only $\*{Z}$ as predictors) and
an unrestricted (full) model, $M_1$, that allows for dependence (e.g., a model for $Y$ using both $X$ and $\*{Z}$).

The fit of each model is assessed by the log-likelihood of the observed data $\*D=\{\left(y_i,\*{z}_i\right)\}_{i=1}^n$
under a parameter vector $\theta$ and its corresponding conditional density $P_\theta$:
\begin{equation}
	\ell_{\*D}(\theta) = \sum_{i=1}^{n} \log P_\theta(y_i|\*{z}_i).
	\label{eq:likelihood}
\end{equation}
Let $\theta_\text{restricted}$ and $\theta_\text{unrestricted}$ denote the maximum-likelihood estimates under the restricted and unrestricted models, respectively.
Under the null hypothesis that the restricted model is correct
the LRT statistic, defined as
\begin{equation}
	T = -2\left(\ell(\theta_\text{restricted})-\ell(\theta_\text{unrestricted})\right),
	\label{eq:lrt test stat}
\end{equation}
asymptotically follows a $\chi^2$ distribution, with degrees of freedom equal to the difference in dimensionality
between the model parameters $\theta_\text{restricted}$ and $\theta_\text{unrestricted}$.
In this context, $T$ measures whether adding $X$ as a predictor provides a statistically significant improvement in model fit.

A key advantage of this approach is its versatility.
As long as the likelihood of the data can be computed
under a chosen class of models, an LRT can be constructed.
This makes the framework particularly well-suited for mixed-data scenarios
when paired with flexible models such as GLMs, 
which can naturally handle a variety of data types.

\vspace{0.4em}
\noindent\textbf{Symmetrical p-values.}
When using LRTs to assess conditional independence between variables, two null
hypotheses can be formulated, each capturing the same independence relationship.
For example, to test whether $X \indep Y \mid \*{Z}$,
the null hypothesis $H_0^Y: P(Y|X,\*{Z})=P(Y|\*{Z})$ is
theoretically equivalent to $H_0^X: P(X|Y,\*{Z})=P(X|\*{Z})$.
However, in finite-sample settings and on mixed data types,
the $p$-values obtained from testing these two hypotheses may differ.
To derive a single $p$-value, $p$, that reflects both results,
Tsgaris et al.~\cite{Tsagris2018} proposed
combining the $p$-values corresponding to $H_0^Y$ and $H_0^X$, denoted $p^Y$ and $p^X$,
as follows:
\begin{equation}
	p = \min\{2\min(p^Y,p^X), \max(p^Y, p^X)\}.
\end{equation}

\subsection{Federated Generalized Linear Models}
\label{sec:background federated glm}
GLMs~\cite{McCullagh1984} are a flexible class of models that generalize ordinary linear regression
by allowing the response variable to follow any distribution from the exponential family.
By specifying a link function that maps the mean of the response distribution to
the linear predictor, GLMs can accommodate a wide range of outcome types, including
continuous, binary, categorical, and ordered. 
This versatility makes GLMs exceptionally powerful for modeling datasets with mixed data types.

\vspace{0.4em}
\noindent \textbf{Maximum Likelihood Estimation (MLE) via IRLS.}
GLM parameters are typically obtained via MLE~\cite{Fisher1922, Fisher1925a}.
However, as the likelihood equations 
are often nonlinear,
approximations and iterative optimization procedures are required.
Iteratively Reweighted Least Squares (IRLS)~\cite{Green1984} provides an efficient
framework for this, encompassing classical approaches such as
Newton-Raphson~\cite{Ypma1995, Osborne1992} and Fisher scoring~\cite{Fisher1925a, Osborne1992}.

Let $X \in \mathbb{R}^{n \times k}$ denote the design matrix for $n$ observations
and $k$ regression terms.
Let $y \in \mathbb{R}^{n}$ be the response vector.
In GLMs, the conditional mean $\mu = \mathbb{E}\left[Y | X\right] \in\mathbb{R}^{n}$ is a
function of the linear predictor $\eta = X\beta \in\mathbb{R}^{n}$,
i.e., $\mu(\eta) = g^{-1}(\eta)$,
where $g(\cdot)$ is a differentiable link function.

Estimation of $\beta$ proceeds via IRLS, which constructs a
sequence $\{\beta^{(t)}\}$, for $t=0, 1, \ldots$,
until convergence to the ML estimate $\beta_{\text{MLE}}$.
For $t = 0$, 
$\beta^{(0)}$ is typically chosen arbitrarily (e.g., zeros).
At iteration $t$, define the linear predictors
$\eta^{(t)} = X\beta^{(t)} \in \mathbb{R}^{n}$ and the mean vector
$\mu^{(t)} = g^{-1}(\eta^{(t)}) \in \mathbb{R}^{n}$.
Because the inverse link can be nonlinear, IRLS applies a
first-order (linear) Taylor approximation of $\mu(\eta)$ around $\eta^{(t)}$:
\begin{equation}
	\label{eq:eta_taylor}
	\mu(\eta) \approx \mu^{(t)} + J^{(t)}(\eta - \eta^{(t)}),
\end{equation}
in which $J^{(t)} \in \mathbb{R}^{n \times n}$ is
the Jacobian for $\mu$ with regards to $\eta$, evaluated at $\eta^{(t)}$.

Rearranging Equation \eqref{eq:eta_taylor} for $\eta$ and replacing the unknown mean $\mu$ with
the observed response vector $y$ defines the current
\textit{working response vector} $z^{(t)} \in \mathbb{R}^{n}$:
\begin{equation}
	z^{(t)} \equiv \eta^{(t)} + (J^{(t)})^{-1}(y - \mu^{(t)}) \approx \eta.
\end{equation}
For an exponential-family response $y_i$,
$\mathrm{Var}(y_i) = V(\mu_i)$,
where $V(\cdot)$ is the model-specific variance function.
Define the diagonal \textit{weight matrix}
$ W^{(t)} \in \mathbb{R}^{n \times n}$ with entries
\begin{equation}
	w_i^{(t)} = \frac{1}{V(\mu_i^{(t)})} \left( \frac{d\mu_i}{d\eta_i}\bigg|_{\eta_i^{(t)}} \right)^{2}.
\end{equation}

Using these quantities, the IRLS update for the coefficient vector is obtained as
the weighted least squares estimator
\begin{equation}
	\beta^{(t+1)} = (X^\top W^{(t)} X)^{-1} X^\top W^{(t)} z^{(t)}.
\end{equation}

To increase the numerical stability of IRLS, several approaches can be used, including Ridge regression~\cite{Hoerl1970}, line search~\cite{Cohen1981}, and Levenberg–Marquardt damping~\cite{Gavin2016}.
Details are provided in Appendix~\ref{sec:appx irls conv}.

\vspace{0.4em}
\noindent\textbf{Federated Model Fitting.}
For the federated computation of the IRLS updates across $k = 1, \ldots, K$ clients,
let $X_k$, $W_k^{(t)}$, and $z_k^{(t)}$ denote the site-specific
design matrix, weight matrix, and working response vector, respectively,
associated with the current global parameter $\beta^{(t)}$.
Since $W^{(t)}$ is diagonal, both the Fisher information matrix $X^\top W^{(t)} X\in\mathbb{R}^{k\times k}$
and the score vector $X^\top W^{(t)} z^{(t)}\in\mathbb{R}^{k}$ can be decomposed as sums of
locally computable statistics~\cite{Chen2013}:
\begin{equation}
	\begin{aligned}
		X^\top W^{(t)} X       & = \sum_{k=1}^K X_k^\top W_k^{(t)} X_k        \\
		X^\top W^{(t)} z^{(t)} & = \sum_{k=1}^K X_k^\top W_k^{(t)} z_k^{(t)}.
	\end{aligned}
	\label{eq:IRLS decomp}
\end{equation}

Chen et al.~\cite{Chen2013} originally introduced this decomposition to parallelize
computations across multiple machines and thereby accelerate model fitting.
However, this additive structure naturally supports a privacy-preserving federated workflow.
Specifically, each client computes its local contributions to the
Fisher information and score vector as in \eqref{eq:IRLS decomp}.
Without sharing any raw data, a central server then aggregates these local quantities
and updates the global parameter vector according to the federated IRLS formula:\looseness=-1
\begin{equation}
	\beta^{(t+1)}=\left( \sum_{k=1}^{K} X_k^\top W_k^{(t)} X_k\right)^{-1} \sum_{k=1}^{K} X_k^\top W_k^{(t)} z_k^{(t)}.
	\label{eq:IRLS fed}
\end{equation}

This iterative procedure converges to the global maximum-likelihood estimator $\beta_{\text{MLE}}$
while providing a strong layer of data protection, since only aggregated statistics 
are shared.

\subsection{Causal Discovery}
Causal discovery is the task of inferring causal relationships,
typically from observational data alone.
In the presence of latent confounding, non-parametric algorithms, which
make no functional or distributional assumptions,
aim to learn a PAG representing the class of
all causal graphs that entail the same set of conditional independencies, known as the
Markov Equivalence Class (MEC)~\cite{Zhang2007}.

Constraint-based causal discovery algorithms infer causal relationships by
exploiting patterns of conditional dependence and independence among observed variables.
A prominent example is the Fast Causal Inference (FCI) algorithm~\cite{Spirtes2001, Zhang2008},
which is sound and complete even under latent confounding and selection bias.
These algorithms typically start from a fully connected graph and,
based on CI tests results, iteratively remove edges and apply orientation rules.

Crucially, most algorithms strongly rely on the faithfulness assumption, which is
often violated in real-world datasets.
As a result, considerable effort has been devoted to improving reliability
by enhancing robustness to unfaithfulness~\cite{Ribeiro2025},
developing more principled and flexible CI tests,
and integrating multiple datasets to increase effective sample size.

\subsection{Integration of Overlapping Datasets (IOD) Algorithm}
\label{sec:background iod}
IOD extends the FCI to handle multiple datasets with non-identical sets of variables, while still accounting for latent confounding.
It relies solely on the faithfulness assumption, under which IOD is provably sound and complete.

Crucially, IOD determines CI relations using Fisher’s method~\cite{Fisher1925}.
For each CI query, all datasets in which the
relevant variables are jointly observed contribute their local test results,
which are subsequently combined to produce a single, aggregated $p$-value.
Formally, if $\{p_i\}_{i=1}^{K}$ are the $p$-values from $K$ independent
tests of the same CI, Fisher’s method combines these $p$-values
by computing the test statistic
\begin{equation}
	T=-2\sum_{i=1}^{K} \log\left(p_i\right),
\end{equation}
which, under the null hypothesis, follows a $\chi^2$-distribution with $2K$ degrees of freedom.
The resulting statistic is then used to obtain a single, combined $p$-value for
a joint decision regarding the conditional independence.

Based on these decisions, the algorithm first infers,
for each locally observed set of variables, the skeleton of the corresponding local graph
and orients unshielded triples according to Rule 0 of the FCI algorithm (see Appendix~\ref{sec:appx fci orient}).
IOD then proceeds by integrating such local structural information
to construct a partially oriented graph over the full set of observed variables.
Because the datasets may differ in variable coverage, some CI
information required to determine the existence of a particular edge may be unavailable.
In such cases, IOD generates multiple candidate graphs representing all admissible alternatives.
Then, the algorithm applies all remaining FCI orientation rules to each candidate graph.
Again, missing CI information may prevent the definitive orientation of certain triples,
either unshielded triples (via Rule 0) or those involved in discriminating paths (via Rule 4)~\cite{Zhang2008}.
Here, the algorithm considers both possible orientations (collider and non-collider),
thereby further expanding the list of candidate graphs.\looseness=-1

Remarkably, this enumeration process can produce candidate graphs that are inconsistent with
some of the observed conditional independencies or that violate the formal
properties of PAGs. IOD therefore performs a final validation step, discarding any graphs that
conflict with the available evidence or with PAG semantics.
The final output is a complete list of PAGs that not only satisfy all observed
CI constraints, but also account for ambiguity arising from unavailable information.

%% file: 03methodology.tex
\section{Federated Causal Discovery with Distinct, Mixed Sets of variables and Latent Confounding}
As described in Section~\ref{sec:background iod}, the existing IOD algorithm~\cite{Tillman2011}
is designed to handle both latent confounding and datasets that include non-identical sets of variables.
However, it does not preserve privacy, as it assumes centralized access to all datasets
in order to compute the required statistics.
Notably, IOD requires data only to determine CI relationships,
and its CI test is a configurable component. As a result, it can be readily replaced,
and any improvement in CI testing directly enhances the algorithm's accuracy.

Building on this insight, we introduce fedCI,
a novel federated CI testing framework that handles datasets with
differing sets of variables and mixed data types. We also
present a newly implemented R package for a privacy-preserving adaptation of the
IOD algorithm, which seamlessly
supports federated CI tests.
Together, these form the fedCI-IOD pipeline,
the first federated solution for causal discovery under latent confounding,
with the added advantages of supporting mixed data types,
accounting for site-specific effects, and fully leveraging the available
information of distributed datasets.
In addition, we improved the computational efficiency of the IOD algorithm,
further strengthening its practical applicability.

In Section~\ref{sec:fed_ci}, we introduce fedCI, and in Section~\ref{sec:fed_IOD},
we show how integrating fedCI with IOD yields a flexible and robust federated causal discovery algorithm.

\subsection{Federated Conditional Independence Testing}
\label{sec:fed_ci}
As detailed in Section~\ref{sec:background lrt}, LRTs
can be employed to assess CI relationships among variables.
When the underlying nested models are estimated in a federated manner,
the resulting LRT naturally yields a federated CI test.
In this work, we implement such tests using federated GLMs.

GLMs offer a flexible and statistically principled modeling approach capable of
accommodating diverse data types and capturing complex relationships through
appropriate link functions and basis expansions.
Because GLMs model response variables drawn from the exponential family and allow predictors of arbitrary types,
they are well suited for the mixed datasets frequently encountered in multi-center studies.
Furthermore, GLMs can be efficiently optimized in a federated setting~\cite{Chen2013},
enabling accurate distributed model fitting without sharing raw data.
Given their modeling flexibility, interpretability, and compatibility with established federated optimization schemes,
GLMs provide a robust and practical foundation for our proposed federated CI testing framework.

\vspace{0.4em}
\noindent\textbf{Data Privacy.}
Because fedCI aims to address the complexities
of real-world causal discovery, ensuring data privacy is fundamental. 
While IRLS abstracts from raw data and is more privacy-preserving than simply sharing the data,
its iterative optimization repeatedly leaks information which makes it prone to model inversion attacks,
particularly for small datasets~\cite{Hannun2021}.
Differential privacy is commonly used to mitigate such attacks, but it may affect accuracy~\cite{Dwork2014}.
Since the required noise for differential privacy is determined solely by the
sensitivity of the data distribution and does not account for sample sizes,
small datasets may become infeasible for modeling due to excessive noise.
To provide stronger privacy guarantees in fedCI while preserving accuracy,
we instead use pairwise additive masking, which masks individual client
contributions without altering the total IRLS update~\cite{Bonawitz2016}.

\vspace{0.4em}
\noindent\textbf{Client Heterogeneity.}
Site-specific effects often complicate the direct aggregation of shared variables across datasets.
To address this, fedCI models the site as a fixed effect using a multinomial variable.
While this effectively captures site-specific effects,
the resulting site-specific coefficients $\alpha$ may reveal sensitive information about differences between clients,
even when applying additive masking.

To address these privacy concerns, we provide an alternative privacy-preserving
fitting strategy in which the joint Fisher-scoring optimization is replaced by a
coordinate-ascent procedure, corresponding to an adapted simplified Fisher-scoring scheme~\cite{MaullinSapey2021}.
We refer to this variant as \emph{fedCI–Coordinate Ascent} (\emph{fedCI-CA}).
It relies on the mild and standard assumption of informational independence
between the predictor coefficients $\beta$ and the site-specific effects $\alpha$.
Under this assumption, each $\alpha_{k,\text{MLE}}^{(t)}$ can be calculated locally for client $k$ at each step $t$,
and used to update the global Fisher information and score vector for $\beta^{(t)}$.
In this setting, the score vector formula becomes:\looseness=-1
\begin{equation}
	\beta^{(t+1)} = (X^\top W^{(t)} X)^{-1} X^\top W^{(t)} \left(z^{(t)}-G\alpha_{\text{MLE}}^{(t)}\right),
\end{equation}
where $G$ is a client-indicator matrix. This matrix can be dropped in the federated update formula by decomposing the global calculation into local sums:
\begin{equation}
	X^\top W^{(t)} \left(z^{(t)}-G\alpha_{\text{MLE}}^{(t)}\right) = \sum_{k=1}^K X_k^\top W_k^{(t)} \left(z_k^{(t)}-\alpha_{k,\text{MLE}}^{(t)}\right).
\end{equation}
This yields a federated calculation of the score vector that accounts for site-specific effects fully locally, without revealing $\alpha$ to the central server or other clients.

\vspace{0.4em}
\noindent\textbf{Combining GLMs and LRTs.}
The log-likelihood formula introduced in Section~\ref{sec:background lrt}
can be directly expressed as a sum over client-wise log-likelihoods.
Specifically, let $\*D$ denote the complete dataset, distributed across $K$ partitions
${D_1,\dots,D_K}$ held by different clients.
The log-likelihood over $\*D$ decomposes into the sum of the log-likelihoods
over all the partitions: 
\begin{equation}
	\begin{aligned}
		\ell_{\*D}(\theta) & = \sum_{i=1}^{|\*{D}|} \log P_\theta(y_i|\*{z}_i)                   \\
		                   & = \sum_{k=1}^K\sum_{i=1}^{|D_k|} \log P_\theta(y_{k,i}|\*{z}_{k,i}) 
		                   & = \sum_{k=1}^K \ell_{D_k}(\theta).
	\end{aligned}
	\label{eq:loglikelihood as sum}
\end{equation}
This enables the calculation of the global log-likelihood without exchanging raw data.
Additionally, additive masking is applied to these likelihoods,
so individual contributions cannot be discerned, adding another layer of privacy protection.

Notably, for Gaussian data the log-likelihood
depends on the dispersion of the observations,
which can be estimated inaccurately when computed only locally under client heterogeneity.
Therefore, instead of exchanging log-likelihood values alone, 
clients exchange summary statistics that allow the dispersion to be properly 
accounted for while maintaining strong data protection. 
Details on handling dispersion in Gaussian data are provided in Appendix~\ref{sec:appx dispersion}.

By integrating federated GLMs for distributed log-likelihood computations
with the LRT and Tsagris et al.'s~\cite{Tsagris2018} method for unifying $p$-values,
we obtain a robust federated procedure for CI testing in heterogeneous, mixed-data settings.
The federated design and masked data transmission provide multiple layers of
privacy protection and eliminate the need of data centralization,
while the unified $p$-value mechanism addresses potential inconsistencies
arising from bidirectional testing.
As a result, this integrated framework provides a consistent and reliable
approach for assessing CI,
forming the core of our federated causal discovery framework.

\vspace{0.4em}
\noindent\textbf{Non-Identical Sets of Variables.}
The CI tests described so far assume that all variables are observed in every dataset.
In practice, however, non-identical variable sets are common and need to be accounted for.
This poses a specific challenge for LRTs, which require that the
nested models being compared are fitted on the exact same set of observations.

To address this, our framework adopts a principled approach for each CI test.
For a given test of $X \indep Y \mid \*Z$, every client verifies whether it has the variables required by the unrestricted models, i.e., $\{X, Y\} \cup \*Z$.
Clients that are able to fit the models do so and send their masked IRLS updates to the server,
whereas those clients who cannot fit the models send masked null-contributions (e.g., zero Fisher information).
With this approach, the privacy-preserving additive masking workflow is kept intact, obfuscating
which clients provided information, without affecting the server's results.

This strategy ensures the validity of the LRT by maintaining a constant effective sample
set for both restricted and unrestricted models, while simultaneously maximizing statistical power by including
all available and relevant data.

\vspace{0.4em}
\noindent\textbf{FedCI Example.}
The proposed fedCI test follows a sequence of steps to assess a CI relationship,
such as $X \indep Y \mid Z$, in a privacy-preserving manner across multiple clients.

\begin{figure*}[ht]
	\centering
	\includegraphics[width=\linewidth]{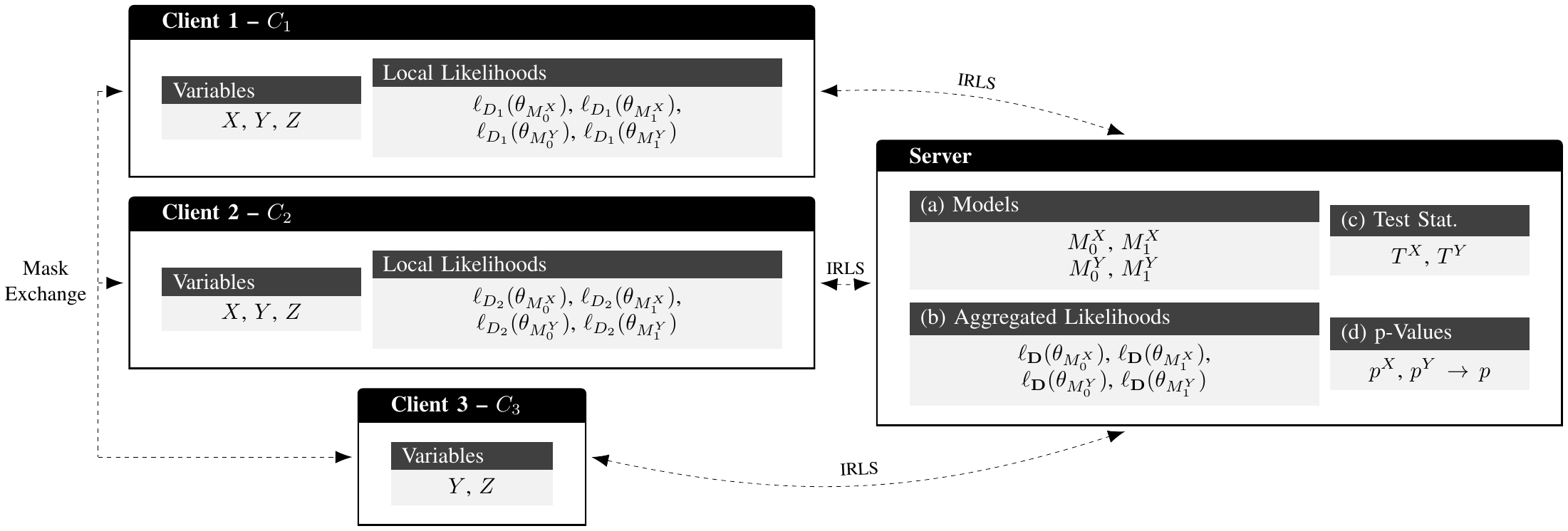}
	\caption{
		Illustration of the federated CI test procedure for $X \indep Y \mid Z$.
		Clients $C_1$ and $C_2$ hold data on variables $\{X, Y, Z\}$, while client $C_3$ only observes $\{Y, Z\}$.
		Since the test requires data on $X$, client $C_3$ cannot contribute, it sends masked null-contributions.
		The server coordinates the federated IRLS fitting procedure to obtain the global models $M_0^X$, $M_1^X$, $M_0^Y$, and $M_1^Y$ and
		then aggregates the local log-likelihoods for each model from the contributing clients to compute test statistics ($T^X, T^Y$),
		$p$-values ($p^X, p^Y$), and a final combined $p$-value ($p$).
	}
	\label{fig:fedci-client-server}
\end{figure*}

To illustrate, consider a setup with three clients as shown in \autoref{fig:fedci-client-server}:
$C_1$ and $C_2$ observe variables $\{X, Y, Z\}$, while $C_3$ only observes the variables $\{Y, Z\}$.

First, the test formulates the bidirectional null hypotheses $H^Y_0$ and $H^X_0$.
For each hypothesis, a pair of nested GLMs is specified: $M^Y_0: Y \sim Z$ and $M^Y_1: Y \sim X, Z$ for $H^Y_0$,
as well as $M^X_0: X \sim Z$ and $M^X_1: X \sim Y, Z$ for $H^X_0$.

The server presents all clients with the test to be evaluated and each model's initial parameter $\beta^{(0)}$.
At each IRLS step, the clients exchange masks in order to perform additive masking and hide their indiviual contributions from the server.
Each client then checks whether it has the required variables --- in this case, $X$, $Y$, and $Z$ --- leaving only $C_1$ and $C_2$ as contributing clients.
Notably, although client $C_3$ cannot fit the models for this particular test,
it still participates in additive masking and returns masked null contributions at each update step,
having no impact on the overall result (see \autoref{eq:IRLS decomp}) while keeping the server unaware.\looseness=-1

The models are then fitted on the contributing clients, $C_1$ and $C_2$, using a federated IRLS algorithm, based on Fisher scoring
(\hyperref[fig:fedci-client-server]{\autoref*{fig:fedci-client-server}a}).
As mentioned in Section~\ref{sec:background federated glm},
this iterative optimization only requires the exchange of aggregate statistics between the clients and a central server.
Due to the masking, the server cannot determine any individual client's contribution to the update step.
After convergence, the server and all contributing clients have agreed on the models' parametrization.
Now, the global log-likelihood for each model is computed by summing the local log-likelihoods from participating clients
as justified by \autoref{eq:loglikelihood as sum}.
This is displayed in \hyperref[fig:fedci-client-server]{\autoref*{fig:fedci-client-server}b},
where the log-likelihood over the set of all datasets $\*D$ is calculated without ever centralizing the datasets $D_1$ and $D_2$.

Finally, as shown in \hyperref[fig:fedci-client-server]{\autoref*{fig:fedci-client-server}c} and
\hyperref[fig:fedci-client-server]{\autoref*{fig:fedci-client-server}d},
the global log-likelihoods are used to compute the LRT statistics ($T^Y, T^X$) and their corresponding $p$-values ($p^Y, p^X$).
These are combined into a single $p$-value using the method from Tsagris et al.~\cite{Tsagris2018} to make a final decision
on the CI hypothesis.

\vspace{0.4em}
\noindent\textbf{FedCI Computation.}
Returning to the introductory example (\autoref{fig:intro example}), we compare the
CI test results obtained using Fisher's method with those of our fedCI test,
both evaluated at a significance level of $0.05$.
In this setup, client $C_1$ has variables $\{A,B,C,E\}$ and $C_2$ has $\{A,C,D,E\}$
As an example, we consider the test for $C \indep E \mid A$, which involves only shared variables
and requires aggregation of local results into a single decision on
whether to reject the null hypothesis.

For the meta-analysis approach, we adopt the same modeling strategy as in the
fedCI test, ensuring that the comparison focuses solely on the
federated aspect, removing differences in model specification as a source of variation.
Thus, each client first performs local CI tests using LRTs based on local GLMs,
following the bidirectional $p$-value combination procedure of~\cite{Tsagris2018}.
Specifically, the local $p$-values $p_1^C$ and $p_1^E$
computed by client $C_1$ are combined into a single $p$-value $p_1$,
and the local $p$-values $p_2^C$ and $p_2^E$ from client $C_2$ are combined into $p_2$.
In total, four models are fit per client, that is the full and nested models
for $H^C_0:P(C|A,E)=P(C|A)$ and $H^E_0:P(E|A,C)=P(E|A)$.
Log-likelihoods, test statistics,
$p$-values, and combined $p$-values for
these fitted regressions are reported in \autoref{table:meta-analysis computation example}.
The client-specific $p$-values $p_1=0.0918$ and $p_2=0.3253$
are then aggregated using Fisher's method,
producing a $\chi^2$-distributed test statistic
with four degrees of freedom $T=-2\left(\ln\left(0.0918\right)+\ln\left(0.3253\right)\right)\approx 7.022$,
which corresponds to a final $p$-value of $0.1347$, leading to a failure to
reject the null hypothesis that $C \indep E \mid A$.

\begin{table}[htbp]
	\caption{Meta-Analysis test example from \autoref{fig:intro example}.}
	\label{table:meta-analysis computation example}
	\centering
	\begin{tabular}{ c|l|c|c|c|c }
		\hline
		Client               & Regression         & Log-Lik.    & Test Stat.                & p                         & Comb. p                   \\
		\hline
		\hline
		\multirow{4}{*}{$1$} & \tiny$C\sim A,1$   & $-16079.38$ & \multirow{2}{*}{$4.7765$} & \multirow{2}{*}{$0.0905$} & \multirow{4}{*}{$0.0918$} \\
		                     & \tiny$C\sim A,E,1$ & $-16076.99$ &                           &                           &                           \\ \cline{2-5}
		                     & \tiny$E\sim A,1$   & $-27065.75$ & \multirow{2}{*}{$4.8054$} & \multirow{2}{*}{$0.0918$} &                           \\
		                     & \tiny$E\sim A,C,1$ & $-27063.34$ &                           &                           &                           \\
		\hline
		\multirow{4}{*}{$2$} & \tiny$C\sim A,1$   & $-16053.22$ & \multirow{2}{*}{$2.2459$} & \multirow{2}{*}{$0.3253$} & \multirow{4}{*}{$0.3253$} \\
		                     & \tiny$C\sim A,E,1$ & $-16052.10$ &                           &                           &                           \\ \cline{2-5}
		                     & \tiny$E\sim A,1$   & $-27095.42$ & \multirow{2}{*}{$2.2734$} & \multirow{2}{*}{$0.3209$} &                           \\
		                     & \tiny$E\sim A,C,1$ & $-27094.28$ &                           &                           &                           \\
		\hline
	\end{tabular}
\end{table}

In contrast, our fedCI test correctly rejects this hypothesis,
indicating that $C \nindep E \mid A$, by leveraging the combined information
from all samples rather than aggregating local results.
The corresponding key metrics are shown in \autoref{table:federated computation example},
highlighting a final $p$-value of $0.0338$.

\begin{table}[htbp]
	\centering
	\caption{FedCI test example from \autoref{fig:intro example}.}
	\label{table:federated computation example}
	\begin{tabular}{ l|c|c|c|c }
		\hline
		Regression         & Log-Lik.    & Test Stat.                & p                         & Comb. p                   \\
		\hline
		\hline
		\tiny$C\sim A,1$   & $-32133.74$ & \multirow{2}{*}{$4.3945$} & \multirow{2}{*}{$0.0338$} & \multirow{4}{*}{$0.0338$} \\
		\tiny$C\sim A,E,1$ & $-32130.36$ &                           &                           &                           \\
		\cline{1-4}
		\tiny$E\sim A,1$   & $-54161.40$ & \multirow{2}{*}{$6.7746$} & \multirow{2}{*}{$0.0327$} &                           \\
		\tiny$E\sim A,C,1$ & $-54157.98$ &                           &                           &                           \\
		\hline
	\end{tabular}
\end{table}

This example shows how local unfaithfulness in client data can lead to erroneous conclusions, even when the aggregated data is faithful.
By utilizing information from all samples,
fedCI is able to fit global models across all clients that more accurately capture the true dependencies between variables.

\vspace{0.4em}
\noindent\textbf{FedCI Package.}
The proposed fedCI test is available at \href{https://imigitlab.uni-muenster.de/CAI4H/fedci}{\textit{imigitlab.uni-muenster.de/CAI4H/fedci}}
and as the Python package \emph{python-fedci} on PyPI.
It is built around two main classes: \texttt{Client} and \texttt{Server}.
A client is initialized with the data loaded as a dataframe.
Variables are automatically interpreted based on their data types:
continuous (float), binary (boolean), ordinal (integer), or multinomial (string).

To operate in the most privacy-preserving manner, the framework requires the
labels of the union of all variables available across participating clients,
as well as the levels of ordinal and multinomial variables.
This should be determined collaboratively in advance and provided to the server during initialization.
Under this configuration, no information about the distribution of variables across individual clients is disclosed
to the server, thereby minimizing information leakage during both setup and execution.

As an alternative, the server may automatically query the client variables to obtain the union of labels and variable levels.
This option simplifies setup but may reveal sensitive information about the distribution of variables across clients,
posing a particular privacy-concern when only a single client can fit a given model.
Accordingly, providing the union of all variables at initialization remains the
recommended configuration for privacy-preserving operation.

Once the server is initialized and connected to all participating clients,
the clients become mutually aware of each other, enabling them to perform
additive masking directly without the involvement of the server.
After this setup, arbitrary CI tests can be executed on the distributed data,
with all network communication implemented via remote procedure calls.
Further implementation details and full documentation are available in the public GitHub repository.

\subsection{Implementation of Federated IOD}
\label{sec:fed_IOD}
Building on fedCI, introduced in Section~\ref{sec:fed_ci}, we now present the adaptations
we made to the IOD algorithm~\cite{Tillman2011} that not only transform it into a fully privacy-preserving
federated framework for multi-center causal discovery but also improve its computational performance.

Under the faithfulness assumption, IOD learns a list of PAGs over the union
of all observed variables, guaranteeing that the true PAG is included even when
datasets differ in their sets of variables and may be affected by latent confounding.
Its original formulation, however, requires centralized access to all datasets to
compute the necessary combined statistics, which compromises data privacy.

In our new R package implementing a privacy-preserving version of IOD,
we ensure that each required CI test is computed locally, and only the
resulting statistics are shared, eliminating the need to centralize raw data.
All subsequent steps of the IOD remain unchanged, ensuring that its theoretical guarantees are preserved.
The CI tests remain configurable, allowing decisions to be made using the most
appropriate modeling for the data. By default, CI test results are combined using
Fisher's method, as originally proposed.

As an alternative, we can use the fedCI test, which assesses CI relations by
directly leveraging all available information across multiple datasets while rigorously protecting data privacy.
This increases statistical power and improves the reliability of the inferred causal relationships.
With these advances, the resulting fedCI-IOD pipeline emerges as the
first federated causal discovery framework capable of handling latent confounding and
heterogeneous datasets with non-identical variable sets, mixed data types,
and site-specific effects.

In addition to replacing IOD's CI testing with a fully federated computation,
we propose an optional modification to the IOD learning process that improves computational efficiency.
In its original formulation, IOD constructs the global PAG over the union of all observed variables
by incorporating orientations implied by unshielded colliders in local PAGs, as
these encode definite non-ancestral relationships that must hold in any valid PAG
containing those variables.
However, this principle applies more broadly:
both ancestral and non-ancestral relationships determined by any triple with order
(colliders and non-colliders triples identified via discriminating paths) in the local PAGs,
are likewise invariant and should be preserved globally.
The original IOD, however, does not exploit these additional orientations.

As a result, IOD may generate PAGs that are inconsistent
with CI constraints implied by the local PAGs,
only to discard them later during the final validation stage.
To address this inefficiency, we extended the algorithm so the global structure
is constructed from the outset incorporating orientations from both
unshielded colliders and all triples with order in the local PAGs.
This prevents the generation of PAGs that would inevitably violate local CI constraints,
thereby reducing the number of candidate structures the algorithm must
process and significantly improving computational efficiency.
Experimental results demonstrating this improvement and a detailed explanation are presented in Appendix~\ref{sec:iod appx}.

\vspace{0.4em}
\noindent\textbf{rIOD Package.}
To our knowledge, no privacy-preserving implementation of the IOD algorithm currently exists,
We provide such an implementation at \href{https://imigitlab.uni-muenster.de/CAI4H/rIOD}{\textit{imigitlab.uni-muenster.de/CAI4H/rIOD}},
supporting both the original approach, which leverages only local invariances from
unshielded colliders to construct the global PAG, and our enhanced version that
leverages invariances from all triples with order (optionally restricted to colliders with order).
Additionally, the package is designed for federated computation, supporting both
the original meta-analysis-based CI testing strategy and the
federated CI testing framework (\textit{fedCI}) introduced in this work.

\vspace{0.4em}
\noindent\textbf{FedCI-IOD WebApp.}
To make the improved and federated IOD algorithm broadly accessible,
we also provide a self-hostable web application that integrates all components introduced in this paper.
The app allows users to upload their own data and perform causal discovery either using
Fisher's method, sharing only $p$-values over the network, or using fedCI,
which performs the federated IRLS procedure using remote procedure calls,
fitting global generalized linear models on all available data while preserving privacy.
The system supports differing sets of variables and mixed data types,
including continuous, binary, ordinal, and multinomial variables.
The implementation is publicly available at \href{https://imigitlab.uni-muenster.de/CAI4H/fedci-iod}{\textit{imigitlab.uni-muenster.de/CAI4H/fedci-iod}}
or as the prebuilt docker image \href{https://hub.docker.com/r/maximilianhahn/fedci-iod-app}{\textit{maximilianhahn/fedci-iod-app}} on docker hub.

%% file: 04simulations.tex
\section{Simulations}

In our simulations, we evaluate the performance and practical benefits of the fedCI-IOD
framework in multi-center settings with heterogeneous datasets and latent confounding.
We consider scenarios with non-identical variable sets, mixed data types, and site-specific effects,
reflecting key challenges in realistic distributed causal discovery applications.

As fedCI-IOD is the first framework designed to operate under these conditions,
existing methods are not directly comparable: most require identical variable sets across datasets,
assume causal sufficiency rather than accommodating latent confounding, and
do not provide deployable implementations suitable for multi-center environments.
Given this, we benchmark our approach against the IOD algorithm with its
original meta-analysis-based CI tests on synthetic datasets.
This comparison isolates the contribution of our fedCI test to the overall
causal discovery process and enables a direct assessment of the resulting gains
in statistical power and structural accuracy.\looseness=-1

In the following, we describe the data generation procedure,
simulation design, and resulting performance evaluation.

\subsection{Data Generation}
\label{sec:data_generation}
As the basis for our simulations, we generate datasets from
$5$-node PAGs that are both horizontally (sample-level) and vertically (feature-level) partitioned.
This setup allows us to evaluate performance under latent confounding and partially overlapping variable sets.
During vertical partitioning, we ensure that three variables are shared
across all partitions while the remaining two are exclusive to different partitions,
reflecting realistic settings in which some centers observe features that others do not.
For example, a dataset with $5,000$ samples for $\{A,B,C,D,E\}$ may be split
into four partitions: two containing $1,250$ samples for $\{A,B,C,D\}$,
and two containing $1,250$ samples for $\{B,C,D,E\}$.
Variables can be continuous, binary, ordinal with four levels, or multinomial with four levels.
To mitigate sparsity in the discrete joint distribution, we require that at least two variables are continuous.
Further, to introduce heterogeneity across sites, we augment these $5$-node PAGs
with a sixth multinomial variable representing the client site, which acts as a parent of all other variables.
This site variable is used to horizontally partition the data, having $4$, $8$, or $12$ levels.
Overall, the tests are performed with total sample sizes $500$, $1,000$, $2,500$, and $5,000$.

Our study utilizes $30$ randomly generated $5$-node PAGs,
selected to ensure structurally challenging scenarios.
One such PAG is the example shown in \autoref{fig:intro example}.
Specifically, each PAG must satisfy two criteria.
First, it must contain at least one collider of order one or higher, thus ensuring
a minimal structural complexity. To guarantee local identifiability of such
colliders, all variables in the associated discriminating path must be
observed together in at least one partition.
Second, each PAG must contain at least three tail connections.
This increases the likelihood of pairs of variables that are conditionally independent
only given a non-empty set of variables.
Accurate causal discovery in such cases
requires correctly identifying not only the conditional independence 
but also the expected conditional dependencies for all proper subsets of the conditioning set.
These scenarios are particularly informative because CI tests often suffer from limited power in small samples.
When causal structures can be recovered using only marginal independencies,
apparent performance may simply reflect structural simplicity rather than reliable statistical evidence.
By focusing on more demanding PAGs, the simulations meaningfully assess CI test reliability for
detecting both independencies and dependencies, and thus the overall causal discovery performance.

Data are generated from these PAGs by first constructing their canonical causal diagram and
then sampling from a conforming Structural Causal Model (SCM)~\cite{Pearl2009, richardson2002ancestral}.
The canonical causal diagram is obtained from a PAG by using the arrowhead augmentation procedure
of~\cite{zhang2006causal}, which converts each $\circ\!\!\!\!\rightarrow$ edge into a
directed edge $\rightarrow$, orients all remaining $\circ\!\!\!-\!\!\!\circ$ edges so that
no unshielded colliders or cycles are created, and replaces any
bidirected edge $A \leftrightarrow B$ by a latent
confounder $U_{AB}$ acting as a common cause of $A$ and $B$. \looseness=-1

In the SCM, each observed variable $X$ is modeled as a function of its
parents within the GLM framework. 
Formally, the linear predictor $\eta$ of $X$ over its parents $Pa(X)$ is defined as\looseness=-1
\begin{equation}
	\eta = \sum_{Pa_i \in Pa(X)} \beta_i Pa_i.
	\label{eq:dgp}
\end{equation}
For continuous variables, $\eta$ is combined with Gaussian noise to
produce $X \sim \mathcal{N}(\eta,1)$.
For binary, ordinal, and multinomial variables, observed values are
sampled from the appropriate distribution with mean
$\mathbb{E}[X] = g^{-1}(\eta)$, where $g^{-1}$ is the inverse link function
chosen according to the variable type.
The coefficients $\beta_i$ for each parent $Pa_i$ are sampled uniformly from $\left[-1, -0.2\right] \cup \left[0.2, 1\right]$
to ensure sufficiently strong causal effects.
This procedure, implemented using the \texttt{simMixedDAG} R package~\cite{lin2019simmixed}, reduces
the likelihood of generating data that is empirically unfaithful to
the conditional dependencies and independencies implied by the underlying canonical causal diagram,
while supporting heterogeneous variable types.
We ran $30$ independent simulations for every configuration of PAG ($30$), sample size ($4$), and client count ($3$),
yielding a total of $10,800$ different simulations or $453,600$ CI tests.

\subsection{Simulation Scenarios}

Utilizing the data generation approach described in Section~\ref{sec:data_generation},
we investigate three simulation scenarios designed to evaluate the viability of the proposed fedCI test.
We compare its performance against two benchmarks:
Fisher's method (meta-analysis) and centralized data analysis (pooled baseline).
These scenarios assess the robustness of the federated approach to variations in sample distribution and its effectiveness
when integrated into the IOD algorithm for causal discovery.
For visual clarity, we first present results for the fedCI variant, with client-site as a shared multinomial variable.
Results for the fedCI-CA variant are provided in Appendix~\ref{sec:appx C}.

The three approaches assess CI relations using GLM-based LRTs.
FedCI fits GLMs in a federated way, leveraging
information across partitions without sharing raw data.
Fisher's method fits GLMs locally within each partition and combines the 
$p$-values to obtain a global test statistic.
Centralized tests serve as a performance baseline by pooling all observations into a single dataset,
ignoring horizontal partitioning and modeling client site as a multinomial variable in each test.

\vspace{0.4em}
\noindent\textbf{Scenario 1: Detection of (In)dependencies.}
For the first simulation, we evaluate the overall ability of the three methods at recovering the (in)dependencies between variables.
Performance is measured by the accuracy of the CI test at determining the true underlying relationship.

We compare the inferred conditional (in)dependencies with the true distributional
relations implied by the underlying causal structure, thereby directly assessing
each method's performance even in the presence of empirical unfaithfulness.

\begin{figure*}[htbp]
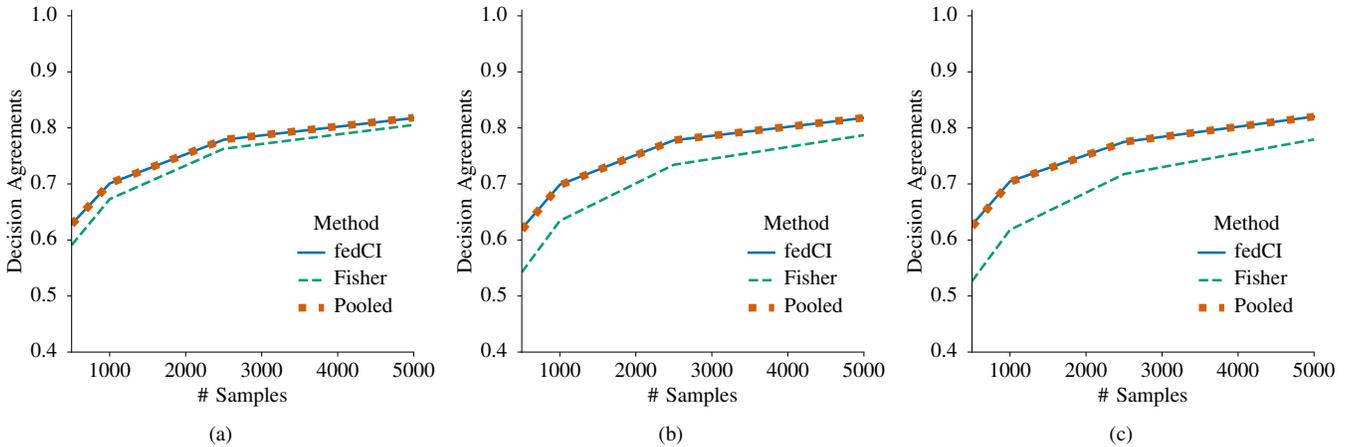

	\centering
	\subfloat[\label{fig:acc comparison 4p}]{\resizebox{0.32\linewidth}{!}{\includesvg[width=0.4\linewidth,inkscapelatex=true]{images/slides_acc/accuracy-all-c4.svg}}}%
	\hfil
	\subfloat[\label{fig:acc comparison 8p}]{\resizebox{0.32\linewidth}{!}{\includesvg[width=0.4\linewidth,inkscapelatex=true]{images/slides_acc/accuracy-all-c8.svg}}}%
	\hfil
	\subfloat[\label{fig:acc comparison 12p}]{\resizebox{0.32\linewidth}{!}{\includesvg[width=0.4\linewidth,inkscapelatex=true]{images/slides_acc/accuracy-all-c12.svg}}}%
	\caption{
		Accuracy of CI tests across (a) $4$, (b) $8$, and (c) $12$ partitions.
		The $y$-axis represents the agreement between test decisions and the true distributional dependence and independence,
		while the $x$-axis denotes the total sample size.
		Results for fedCI (solid line) nearly perfectly align with the pooled baseline (dotted line),
		whereas Fisher's method (dashed line) exhibits performance degradation as the number of partitions increases.
	}
	\label{fig:acc comparison}
\end{figure*}


\autoref{fig:acc comparison} reports the percentage of correctly identified CI relations
as an initial benchmark for each approach.
As expected, accuracy improves with increasing sample size for all methods,
underscoring the importance of reliable federated inference for leveraging larger
effective sample sizes and mitigating the impact of empirical unfaithfulness in small-sample regimes.

Notably, fedCI closely matches the performance of the centralized baseline across
all partitioning levels, with negligible information loss,
demonstrating that federated estimation effectively optimizes the global log-likelihood.
In contrast, Fisher's method consistently underperforms and shows clear sensitivity to
the degree of partitioning, with accuracy deteriorating as the number of partitions increases.
Additional plots providing a more detailed breakdown of accuracy in assessing conditional
(in)dependencies are presented in Appendix~\ref{sec:appx C}.

\vspace{0.4em}
\noindent\textbf{Scenario 2: Difference to Pooled Data.}
The second simulation evaluates how $p$-values from fedCI and Fisher's method deviate
from those obtained using a centralized pooled CI test.
Because the pooled test leverages all available data jointly, it provides the
most statistically powerful benchmark, serving as baseline for comparison.
We quantify deviations by computing 
$\log\left(\frac{p_{\text{dist}}}{p_{\text{pool}}}\right)$, in which
$p_\text{dist}$ denotes the $p$-value from a distributed method and
$p_{\text{pool}}$ the $p$-value from the corresponding pooled CI test.
A log-ratio of zero indicates perfect agreement with the pooled CI test,
while the logarithmic scale ensures comparability across both small and large $p$-values.

\begin{figure}[h]
	\centering
	\resizebox{\linewidth}{!}{
		\includesvg[width=1.3\linewidth,inkscapelatex=true]{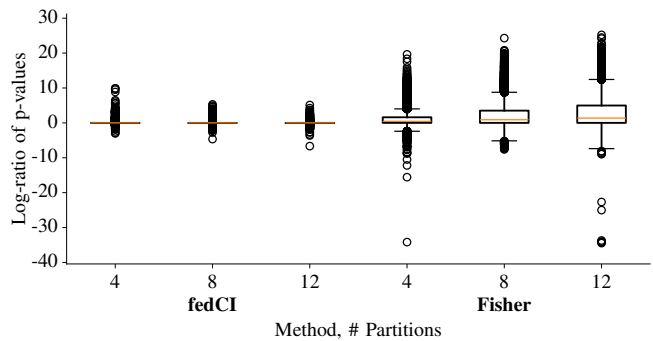}
	}
	\caption{
		Boxplot of log-ratios of $p$-values for fedCI and Fisher's method relative to the pooled baseline on $1,000$ samples, across different numbers of partitions (sites).
		FedCI remains centered at zero with low variance, while Fisher's method increasingly deviates with a positive bias as partitions grows.
	}
	\label{fig:logratio by sample 1000s}
\end{figure}

\autoref{fig:logratio by sample 1000s} displays these results for $1,000$ samples.
FedCI produces $p$-values mostly indistinguishable from the pooled baseline,
demonstrating strong robustness to partitioning, regardless of the chosen significance level.
In contrast, Fisher's method shows overall higher variance and often produces positive log-ratios,
reflecting a conservative bias toward larger $p$-values.
Results for other sample sizes are provided in Appendix~\ref{sec:appx C},
where the same pattern is observed.
In constraint-based causal discovery, where failure to reject the null hypothesis is
taken as evidence of independence,
this conservative bias is critical, at it increases the risk of type II errors and,
in turn, incorrect causal conclusions.\looseness=-1

We also evaluated the fedCI-CA variant, which preserves privacy of client-site effects.
Its performance closely matches that of the fedCI with Fisher-scoring optimization.
\autoref{fig:logratio by sample 1000s CA} shows the corresponding log-ratio boxplots,
confirming that both optimization strategies achieve comparable accuracy.

\vspace{0.4em}
\noindent\textbf{Scenario 3: Causal Discovery.}
Finally, we evaluate how the different CI test approaches affect causal discovery when used within the IOD algorithm.
As a reference, we define the ground truth as the list of PAGs obtained by
running IOD with an oracle CI test under the same variable-overlap constraints as in the simulation.
Here, the oracle CI test returns the true CI relations implied by the
underlying canonical causal diagrams, marginalized to the variables observed in each local dataset.

Performance in causal discovery is typically evaluated using the
Structural Hamming Distance (SHD) between the ground-truth and inferred PAGs~\cite{Jabbari2017},
which counts differences in adjacencies and edge marks.
For better interpretability, we normalize SHD by the maximum possible distance ($20$ for $5$-node PAGs).
Since IOD outputs a list of PAGs rather than a single graph, evaluation must be defined at the list level.
Thus, for each method, we compare its inferred list of PAGs with the reference list by
computing normalized SHDs for all pairwise combinations of PAGs across the two lists and
reporting the minimum (best) value.
This \emph{best normalized SHD} metric reflects the closest agreement between an inferred and reference PAG,
thereby capturing accuracy while
avoiding the need for explicit one-to-one matching between the two lists, which may differ
in length and composition.

As shown in \autoref{table:shd}, the fedCI-IOD approach produces PAGs that are, on average,
nearly identical to those obtained with the centralized pooled CI test,
whereas IOD with Fisher's method produces PAGs with larger SHD values.\looseness=-1

\begin{table}[htbp]
	\centering
	\caption{
		Best normalized SHDs averaged across simulation runs, stratified by
		sample size and number of partitions.
	}
	\label{table:shd}
	\begin{tabular}{ l|r|r|r|r|r }
		\hline
		\multirow{2}{*}{Algorithm} & \multirow{2}{*}{Part.} & \multicolumn{4}{c}{Samples}                               \\
		                           &                        & $500$                       & $1,000$ & $2,500$ & $5,000$ \\
		\hline
		\hline
		\multirow{3}{*}{Pooled}    & $4$                    & $0.337$                     & $0.286$ & $0.217$ & $0.174$ \\
		                           & $8$                    & $0.348$                     & $0.289$ & $0.224$ & $0.182$ \\
		                           & $12$                   & $0.344$                     & $0.287$ & $0.213$ & $0.166$ \\
		\hline
		\multirow{3}{*}{Fisher}    & $4$                    & $0.353$                     & $0.306$ & $0.230$ & $0.190$ \\
		                           & $8$                    & $0.393$                     & $0.334$ & $0.258$ & $0.216$ \\
		                           & $12$                   & $0.403$                     & $0.344$ & $0.268$ & $0.208$ \\
		\hline
		\multirow{3}{*}{fedCI}     & $4$                    & $0.335$                     & $0.287$ & $0.216$ & $0.175$ \\
		                           & $8$                    & $0.346$                     & $0.289$ & $0.224$ & $0.181$ \\
		                           & $12$                   & $0.341$                     & $0.286$ & $0.212$ & $0.166$ \\
		\hline
	\end{tabular}
\end{table}

Since \autoref{table:shd} does not convey the magnitude of differences in
best normalized SHDs across methods,
\autoref{table:shd diff} reports these differences using Cohen's $d$, defined as
$d = \bar{x}_D / s_D$,
where $\bar{x}_D$ is the 
average difference relative to the IOD with pooled CI test
across all $30$ simulations for the same PAGs, sample sizes, and partitions,
and $s_D$ is the corresponding standard deviation. 
This provides a standardized measure of deviation from the IOD with pooled CI test,
with negative values indicating better and positive values worse accuracy.

\begin{table}[htbp]
	\centering
	\caption{
		Cohen's $d$ for the differences in best normalized SHDs between the pooled PAGs
		and the PAGs inferred using Fisher's method or fedCI, stratified by
		sample size and number of partitions.
	}
	\label{table:shd diff}
	\begin{tabular}{ l|r|r|r|r|r }
		\hline

		\multirow{2}{*}{Algorithm} & \multirow{2}{*}{Part.} & \multicolumn{4}{c}{Samples}                                  \\
		                           &                        & $500$                       & $1,000$  & $2,500$  & $5,000$  \\
		\hline
		\hline
		\multirow{3}{*}{Fisher}    & $4$                    & $0.159$                     & $0.162$  & $0.103$  & $0.171$  \\
		                           & $8$                    & $0.378$                     & $0.364$  & $0.223$  & $0.302$  \\
		                           & $12$                   & $0.455$                     & $0.366$  & $0.427$  & $0.411$  \\
		\hline
		\multirow{3}{*}{fedCI}     & $4$                    & $-0.024$                    & $-0.053$ & $-0.033$ & $0.090$  \\
		                           & $8$                    & $-0.047$                    & $-0.026$ & $0.039$  & $-0.056$ \\
		                           & $12$                   & $-0.097$                    & $-0.018$ & $-0.053$ & $-0.055$ \\
		\hline
	\end{tabular}
\end{table}

The results further confirm that fedCI-IOD closely matches the pooled baseline,
while IOD with Fisher's method deviates substantially, indicating that
federated CI testing preserves causal discovery accuracy, whereas
the reduced power and conservative bias of Fisher's method translate into
structural errors in the inferred PAGs.

%% file: 05conclusion.tex
\section{Conclusion, Limitations, and Future Work}

This paper introduced fedCI and its integration with the IOD algorithm as a powerful
solution for federated causal discovery across heterogeneous datasets in distributed environments.
Unlike existing methods, our framework, fedCI-IOD, uniquely handles
datasets with non-identical variable sets, mixed data types, site-specific effects,
and latent confounding, all while rigorously preserving data privacy.
This represents a substantial methodological innovation,
as no prior approach simultaneously addresses these challenges.

Our empirical results demonstrate that fedCI effectively
aggregates information across distributed sites, achieving a level of
statistical power comparable to a centralized analysis on pooled data.
This capability translates into enhanced robustness to local unfaithfulness and
substantially more accurate CI assessments,
often correctly identifying dependencies that traditional meta-analysis approaches,
such as Fisher's method, fail to detect.
Crucially, when integrated into IOD, the resulting fedCI-IOD framework
produces more reliable PAGs, enabling robust, privacy-preserving
causal discovery in realistic, distributed data settings.\looseness=-1

Beyond methodological contributions, we provide a fully deployable software ecosystem,
comprising a Python package for fedCI,
a privacy-preserving R implementation of the IOD algorithm, and
a self-hostable web application for the fedCI-IOD framework.
By making these tools open-source and publicly available, we aim to support the research community,
promote reproducibility, and facilitate practical applications in real-world multi-center studies.

The current implementation of fedCI relies on a
GLM-based modeling framework.
While this imposes linearity in the parameters, potentially limiting the representation
of highly complex relationships, it was a deliberate and advantageous choice:
GLMs not only provide strong statistical guarantees, interpretability, and computational stability,
but also establish a flexible foundation for future extensions.
Currently, fedCI models site variability as fixed effects, but
federated Generalized Linear Mixed Models (GLMMs) could naturally extend the
framework to treat sites as random effects, which is particularly beneficial when
scaling to many clients or when sites are viewed as samples
from a broader population of data-generating environments.
Beyond scalability, GLMMs offer a principled approach to modeling
hierarchical structure and correlated observations, further enhancing
fedCI's ability to support causal discovery in non-i.i.d.
distributed settings~\cite{Ribeiro2020}.

A key strength of the fedCI-IOD framework is its modular architecture.
Federated model estimation, LRT for CI testing, and the privacy-preserving
IOD procedure for causal discovery are implemented as distinct components,
allowing each part to be extended or replaced without modifying the overall pipeline.
This design provides a flexible foundation for future developments.
Advancing these directions, particularly by incorporating federated GLMMs or
more flexible non-parametric approaches in the model estimation stage,
will broaden the applicability of fedCI-IOD,
while preserving its core guarantees of privacy-preserving, distributed causal discovery.

%% file: 99appendix_A.tex
\appendices
\section{IRLS Convergence}
\label{sec:appx irls conv}
IRLS faces convergence issues, especially for non-continuous response variables.
Perfect separation and collinearity in predictors are common causes for diverging coefficients and malconditioned Hessians,
leading to convergence issues.
Therefore, three key mechanisms are added to the base IRLS update with fixed effects:
ridge regularization~\cite{Hoerl1970}, line search~\cite{Cohen1981}, and Levenberg-Marquardt (LM) damping~\cite{Gavin2016}.
With these added mechanisms, convergence significantly improves.
In the following, the mechanisms and their use-case are explained in more detail.

\subsection{Ridge Regularization}
Ridge~\cite{Hoerl1970} is a L2 regularization technique which incorperates the norm of the coefficient vector into the likelihood objective,
so that optimizing
\begin{equation*}
	\ell_{ridge}\left(\beta\right)=\ell\left(\beta\right) + \frac{\lambda_{ridge}}{2}\beta^\top\beta
\end{equation*}
balances the data fit $\ell\left(\beta\right)$ with the size of $\beta$ according to the ridge parameter $\lambda_{ridge} \ge 0$,
where a larger $\lambda_{ridge}$ punishes larger $\beta$ values more strongly.
This regularization is especially useful when encountering strong collinearity within the data.
Typically, $\lambda_{ridge}$ is subject to hyperparameter optimization.

\subsection{Line Search}
Since the true likelihood equation is locally approximated in IRLS,
an update step may have the correct descent direction, though the step size may be so large that it overshoots the minimum,
such that the new $\beta^{(t+1)}$ actually worsens the model fit:\looseness=-1
\begin{equation*}
	\ell\left(\beta^{(t)}\right) > \ell\left(\beta^{(t+1)}\right).
\end{equation*}

Here, line search~\cite{Cohen1981} dampens the update step $\delta^{(t)}$ by some coefficient $\lambda_{LS}$ with $0 < \lambda_{LS} \le 1$,
so that each update step yields an improvement in likelihood, updating $\beta$ by calculating
\begin{equation*}
	\beta^{(t+1)} = \beta^{(t)} + \lambda_{LS}\delta^{(t)}.
\end{equation*}

Since the optimal $\lambda_{LS}$ is not known, starting from the full update with $\lambda_{LS}=1$, $\lambda_{LS}$
is repeatedly reduced until it reaches a value so that an improvement is made.
This is computational expensive because it requires repeated calculation of potential update steps.
Practically, $\lambda_{LS}$ is repeatedly halved with a lower bound that stops the IRLS procedure
should no update step be possible with any $\lambda_{LS}>\epsilon$.

\subsection{Levenberg-Marquardt Damping}
Levenberg-Marquardt Damping (LM Damping) counteracts convergence issues due to a plateauing likelihood surface by dynamically adjusting the iterative optimization procedure to behave more like a Gauss-Newton method when the Fisher Information
is well-behaved and behaving more like gradient ascent when it is not~\cite{Gavin2016}.

With the new beta update formula
\begin{equation*}
	\beta^{(t+1)} = \left(X^\top WX + \lambda_{LM}I\right)^{-1} X^\top Wz,
\end{equation*}
LM Damping ensures the Hessian matrix $X^\top WX$ is positive definite and therfore ensures invertability.
The parameter $\lambda_{LM} \ge 0$ is dynamic and may change at each iteration, where values close to zero give the original Gauss-Newton IRLS step
and larger values make this more like gradient descent, since the influence of the Hessian becomes negligible.

Whenever an update step succeeds, $\lambda_{LM}$ is reduced, moving towards a faster IRLS convergence, and when the update fails,
it is increased, moving towards a slower but more stable gradient ascent.
Success and failure can be determined in different ways, though typically it is determined by whether actual improvement matches the predicted improvement,
reducing $\lambda_{LM}$ whenever the new likelihood is similar to the prediction and
increasing it whenever the actual likelihood is worse than the prediction~\cite{Gavin2016}.
Simpler approaches increase and decrease $\lambda_{LM}$ depending on whether or not the calculated update step increased the likelihood.

%% file: 99appendix_B.tex
\section{FCI Orientation Rules}
\label{sec:appx fci orient}

The original FCI algorithm introduced in \cite{Spirtes2001} consisted of five orientation rules $\mathcal{R}0-\mathcal{R}4$, which was shown to be only complete for arrowhead orientation.
This set of rules was extended by \cite{Zhang2008} with six additional rules $\mathcal{R}5-\mathcal{R}10$,
completing tail orientations and identification of selection bias (undirected edges),
thus achieving full completeness:
with perfect CI information, the algorithm recovers the maximally informative PAG corresponding to the MEC of the true causal graph.\looseness=-1

The FCI algorithm and the complete set of orientation rules is formulated by \cite{Zhang2008} as:
\begin{itemize}
	\item[F1] Form a complete graph $\mathcal{U}$ on the set of variables, in which there is an edge $\circ\!\!-\!\!\circ$ between every pair of variables;
	\item[F2] For every pair of variables $\alpha$ and $\beta$, search \emph{in some clever way} for a set of other variables that render the two independent.
	      If such a set $\*S$ is found, remove the edge between $\alpha$ and $\beta$ in $\mathcal{U}$, and record $\*S$ as $Sepset(\alpha, \beta)$;
	\item[F3] Let $\mathcal{P}$ be the graph resulting from step F2. Execute the orientation rule:
	      \begin{itemize}
		      \item[$\mathcal{R}0$] For each unshielded triple $\langle \alpha, \gamma, \beta \rangle$ in $\mathcal{P}$,
		            orient it as a collider $\alpha \ast\!\!\!\rightarrow\gamma\leftarrow\!\!\!\ast\beta$ if and only if $\gamma$ is not in $Sepset(\alpha, \beta)$.
	      \end{itemize}
	\item[F4] Execute the following mark inference rules until none of them applies:
	      \begin{itemize}
		      \item[$\mathcal{R}1$] If $\alpha\ast\!\!\!\rightarrow\beta\circ\!\!\!-\!\!\!\ast\gamma$, and $\alpha$ and $\gamma$ are not adjacent, then orient the triple as $\alpha\ast\!\!\!\rightarrow\beta\rightarrow\gamma$. 
		      \item[$\mathcal{R}2$] If $\alpha\rightarrow\beta\ast\!\!\!\rightarrow\gamma$ or $\alpha\ast\!\!\!\rightarrow\beta\rightarrow\gamma$, and $\alpha\ast\!\!\!-\!\!\!\circ\gamma$,
		            then orient $\alpha\ast\!\!\!-\!\!\!\circ\gamma$ as $\alpha\ast\!\!\!\rightarrow\gamma$. 
		      \item[$\mathcal{R}3$] If $\alpha\ast\!\!\!\rightarrow\beta\leftarrow\!\!\!\ast\gamma$, $\alpha\ast\!\!\!-\!\!\!\circ\theta\circ\!\!\!-\!\!\!\ast\gamma$,
		            $\alpha$ and $\gamma$ are not adjacent, and $\theta\ast\!\!\!-\!\!\!\circ\beta$, then orient $\theta\ast\!\!\!\rightarrow\beta$. 
		      \item[$\mathcal{R}4$] If $u=\langle\theta,\dots,\alpha,\beta,\gamma\rangle$ is a discriminating path between $\theta$ and $\gamma$ for $\beta$, and $\beta\circ\!\!\!-\!\!\!\ast\gamma$;
		            then if $\beta\in Sepset(\theta, \gamma)$, orient $\beta\circ\!\!-\!\!\ast\gamma$ as $\beta\rightarrow\gamma$;
		            otherwise orient the triple $\langle\alpha,\beta,\gamma\rangle$ as $\alpha\leftarrow\!\!\!\rightarrow\beta\leftarrow\!\!\!\rightarrow\gamma$.
		      \item[$\mathcal{R}5$] For every (remaining) $\alpha\circ\!\!-\!\!\circ\beta$, if there is an uncovered circle path (a path of only unshielded triples and only of edges with $\circ$-marks)
		            $p=\langle\alpha,\gamma,\dots,\theta,\beta\rangle$ between $\alpha$ and $\beta$ s.t. $\alpha$,$\theta$ are not adjacent and $\beta$, $\gamma$ are not adjacent, then orient $\alpha\circ\!\!\!-\!\!\!\circ\beta$
		            and every edge on $p$ as undirected edges ($-$).
		      \item[$\mathcal{R}6$] If $\alpha-\beta\circ\!-\!\!\!\ast\gamma$ ($\alpha$ and $\gamma$ may or may not be adjacent), then orient $\beta\circ\!\!\!-\!\!\!\ast\gamma$ as $\beta-\!\!\!\ast\gamma$.
		      \item[$\mathcal{R}7$] If $\alpha-\!\!\circ\beta\circ\!\!-\!\!\ast\gamma$, and $\alpha$, $\gamma$ are not adjacent, then orient $\beta\circ\!\!\!-\!\!\!\ast\gamma$ as $\beta-\!\!\!\ast\gamma$.
		      \item[$\mathcal{R}8$] If $\alpha\rightarrow\beta\rightarrow\gamma$ or $\alpha-\!\!\!\circ\beta\rightarrow$, and $\alpha\circ\!\!\!\rightarrow\gamma$, orient $\alpha\circ\!\!\!\rightarrow\gamma$ as $\alpha\rightarrow\gamma$ 
		      \item[$\mathcal{R}9$] If $\alpha\circ\!\!\!\rightarrow\gamma$, and $p=\langle\alpha,\beta,\theta,\dots,\gamma\rangle$ is an uncovered potentially directed path
		            (a path of only unshielded triples that be oriented from $\alpha$ to $\gamma$ with appropriate arrowheads and tails) from $\alpha$ to $\gamma$,
		            such that $\gamma$ and $\beta$ are not adjacent, then orient $\alpha\circ\!\!\!\rightarrow$ as $\alpha\rightarrow\gamma$.
		      \item[$\mathcal{R}10$] Suppose $\alpha\circ\!\!\!\rightarrow\gamma$, $\beta\rightarrow\gamma\leftarrow\theta$, $p_1$ is an uncovered potentially directed path from $\alpha$ to $\beta$,
		            and $p_2$ is an uncovered potentially directed path from $\alpha$ to $\theta$. Let $\mu$ be the vertex adjacent to $\alpha$ on $p_1$ ($\mu$ could be $\beta$), and $\omega$
		            be the vertex adjacent to $\alpha$ on $p_2$ ($\omega$ could be $\theta$).
		            If $\mu$ and $\omega$ are distinct, and are not adjacent, then orient $\alpha\circ\!\!\!\rightarrow\gamma$ as $\alpha\rightarrow\gamma$
	      \end{itemize}
\end{itemize}


%% file: 99appendix_C.tex
\section{Handling Dispersion of Gaussian Data}
\label{sec:appx dispersion}

Since the log-likelihood calculation of Gaussian data involves the dispersion of observations,
local modeling under client heterogeniety yields potentially wrong results.
To address this, the MLE of the variance $\hat{\sigma}^2$ is calculated on the server instead by aggregating
the local sums of squared residuals ($RSS_i$) and number of
contributing samples ($n_i$) from each client $i\in\{1,\dots K\}$, with $n=\sum_{i=1}^{K}n_i$.
Specifically, $\hat{\sigma}^2$ is computed as follows:
\begin{equation}
	\begin{aligned}
		\hat{\sigma}^2 & = \frac{1}{n}\sum_{i=1}^K RSS_i \\
	\end{aligned}
	\label{eq:mle for variance}
\end{equation}
Thus, the Gaussian log-likelihood evaluated at $\sigma=\hat{\sigma}$ is: 
\begin{equation}
	\begin{aligned}
		\ell_{\*D}(\theta) & = -\frac{n}{2}\log \left(2\pi\sigma^2\right)- \frac{1}{2\sigma^2}\sum_{i=1}^{K}RSS_i            \\
		                   & = -\frac{n}{2}\log \left(2\pi\sigma^2\right)- \frac{1}{2\sigma^2}n\hat{\sigma}^2                \\
		                   & \overset{\sigma=\hat{\sigma}}{=} -\frac{n}{2} \left(\log \left(2\pi\sigma^2\right)  + 1\right).
	\end{aligned}
	\label{eq:loglikelihood with rss}
\end{equation}

All the exchanged parameters are obfuscated with additive masking to enhance privacy.
While this approach can be seen as exposing more client information than exchanging only log-likelihoods,
we consider the trade-off worthwhile given the resulting improvements in accuracy under client heterogeneity.

%% file: 99appendix_D.tex
\section{Computational Improvement of the IOD Algorithm}
\label{sec:iod appx}
In Appendix~\ref{sec:descr_improved_iod_app}, we present the motivation
and describe the design of our proposed adaptations to the IOD algorithm
aimed at improving its computational performance. Specifically, we
introduce two modified versions. The first, \emph{collider-with-order},
incorporates non-ancestral relationships derived from all colliders with
order identified in the local PAGs. The second, \emph{triple-with-order},
incorporates both ancestral and non-ancestral relationships derived from
all triples with order in the local PAGs, including both colliders and
non-colliders. Both variants leverage these relationships to reduce
computational cost while preserving the correctness of the inferred
structures. Appendix~\ref{sec:exper_improved_iod_app} then compares the
original IOD procedure with these modified variants and discusses the
implications of these changes for both accuracy and computational
efficiency.

\balance
\subsection{Algorithmic Modifications for Improved Efficiency}
\label{sec:descr_improved_iod_app}
Revisiting the IOD, its objective is to construct the list of all PAGs over the
full set of observed variables that are compatible with the data and therefore,
under faithfulness, includes the true underlying PAG.

It starts by extracting, from each local subset of variables,
the graph skeleton and the non-ancestral relations implied by unshielded colliders.
It then constructs a list of candidate global graphs that integrates this
information and ensures that every feasible configuration of edges that cannot be
resolved due to missing CI information is represented.
Subsequently, the FCI orientation rules are applied to each candidate graph.
For Rule 4, which depends on CI information that may be
unavailable, IOD applies an alternative procedure that generates two
PAGs: one orienting the triple in a discriminating path as a collider and
another as a non-collider. This expansion is warranted only when it is unclear
whether the discriminated variable lies in the separating set of the first and
last variables of the discriminating path. When this information is available
because the variables were jointly observed in a dataset, one of the two PAGs is
immediately inconsistent with the data. As a result, some PAGs are produced
only to be discarded during validation, unnecessarily increasing computational
effort.

To make the construction of plausible PAGs more efficient, we propose fully
orienting the local PAGs and further integrating (non-)ancestral relationships
derived from triples with order into the global PAGs.

It is well established that, in addition to the skeleton and unshielded colliders,
all colliders with order are required to fully characterize the MEC of a
PAG~\cite{ali2009markov}. This implies that the non-adjacencies
and (non-)ancestral relations they encode are invariances that must persist in the
global PAG, and that all remaining orientations can be deduced solely through the
FCI rules without additional CI information. The original IOD
did not incorporate orientations based on colliders with order, likely because
identifying the discriminating paths that define them is computationally challenging.

More recently, however,~\cite{claassen2022greedy} introduced an algorithm that
recursively and efficiently identifies all triples with order, including both
colliders and non-colliders, and demonstrated that the MEC of a PAG can
alternatively be fully characterized by its skeleton together with all triples
with order. This enables the systematic extraction of the (non-)ancestral
relations implied by these triples. Incorporating this information into the global
PAGs allows us to efficiently exploit key learned invariances among the variables,
substantially reducing ambiguity in the inferred structures and preventing the
generation of PAGs that contradict the data. Consequently, fewer incorrect
candidate PAGs must be constructed and validated, yielding notable computational
efficiency gains.

\subsection{Experimental Evaluation}
\label{sec:exper_improved_iod_app}
To assess the efficiency gains of the proposed IOD adaptations, we compare the number of
PAGs generated before they are discarded due to violations of local constraints. We evaluate
three variants: the original IOD; the collider-with-order version, which additionally
incorporates non-ancestral relationships from local colliders with order; and the triple-with-order
version, which incorporates ancestral and non-ancestral relationships
from all local triples with order, including both
colliders and non-colliders. An oracle providing the correct CI
relations implied by the true model is used, consistent with the algorithm's assumption of
faithfulness.

We generated 100 random 5-node PAGs over $\mathbf{V} = \{A,B,C,D,E\}$, ensuring that
each contained at least one collider of order one or higher. For each simulation, we selected
two subsets of 4 variables 
from $\mathbf{V}$. Subset selection prioritized variables from discriminating paths.
If only one path was present, the second subset was chosen randomly,
ensuring an overlap of at least three variables. Given the subset size of 4,
the highest observable order for triples in the local PAG is 1. All
three IOD versions were then executed using the oracle CI Test.

\begin{figure}[tb!]
	\centering
	\subfloat[CWO vs. original]{\includegraphics[width=0.32\linewidth]{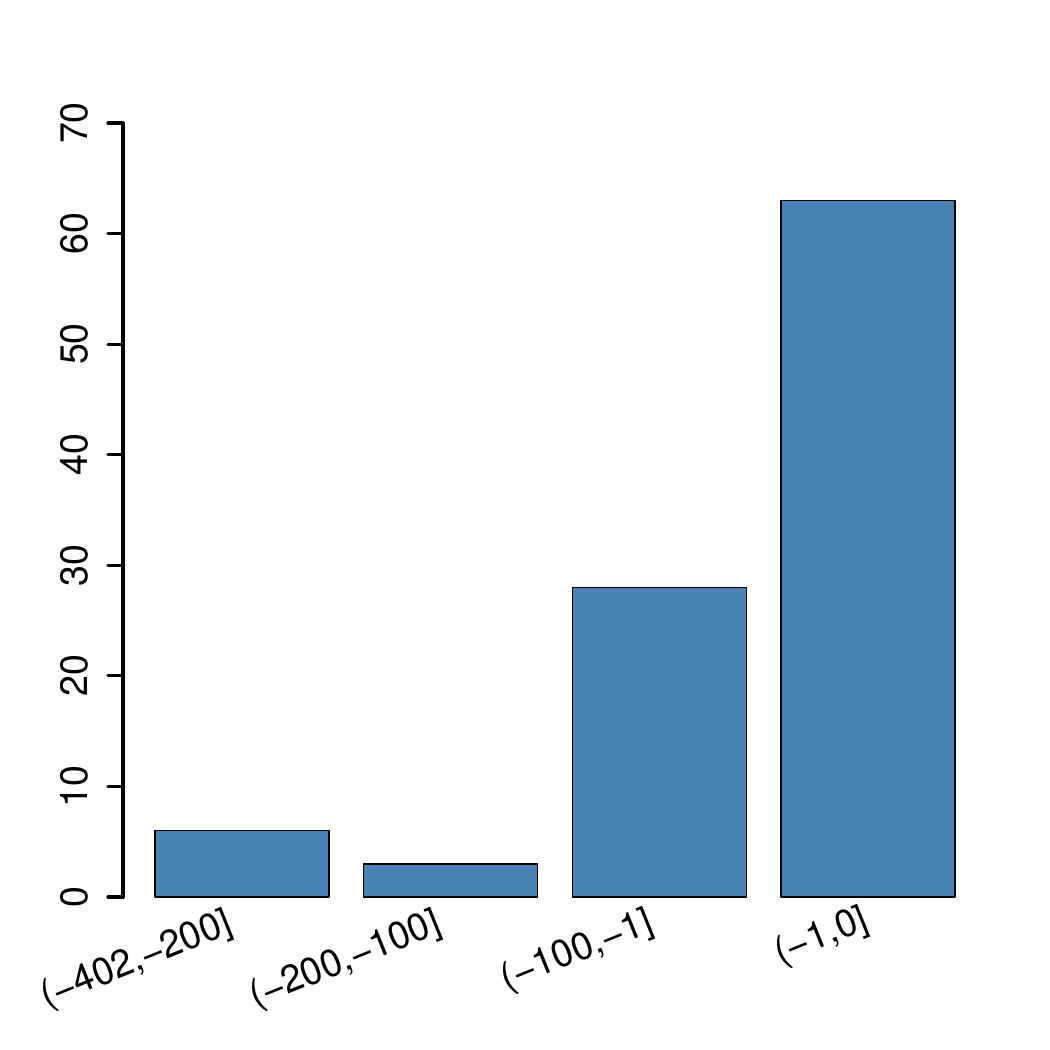}}%
	\subfloat[TWO vs. original]{\includegraphics[width=0.32\linewidth]{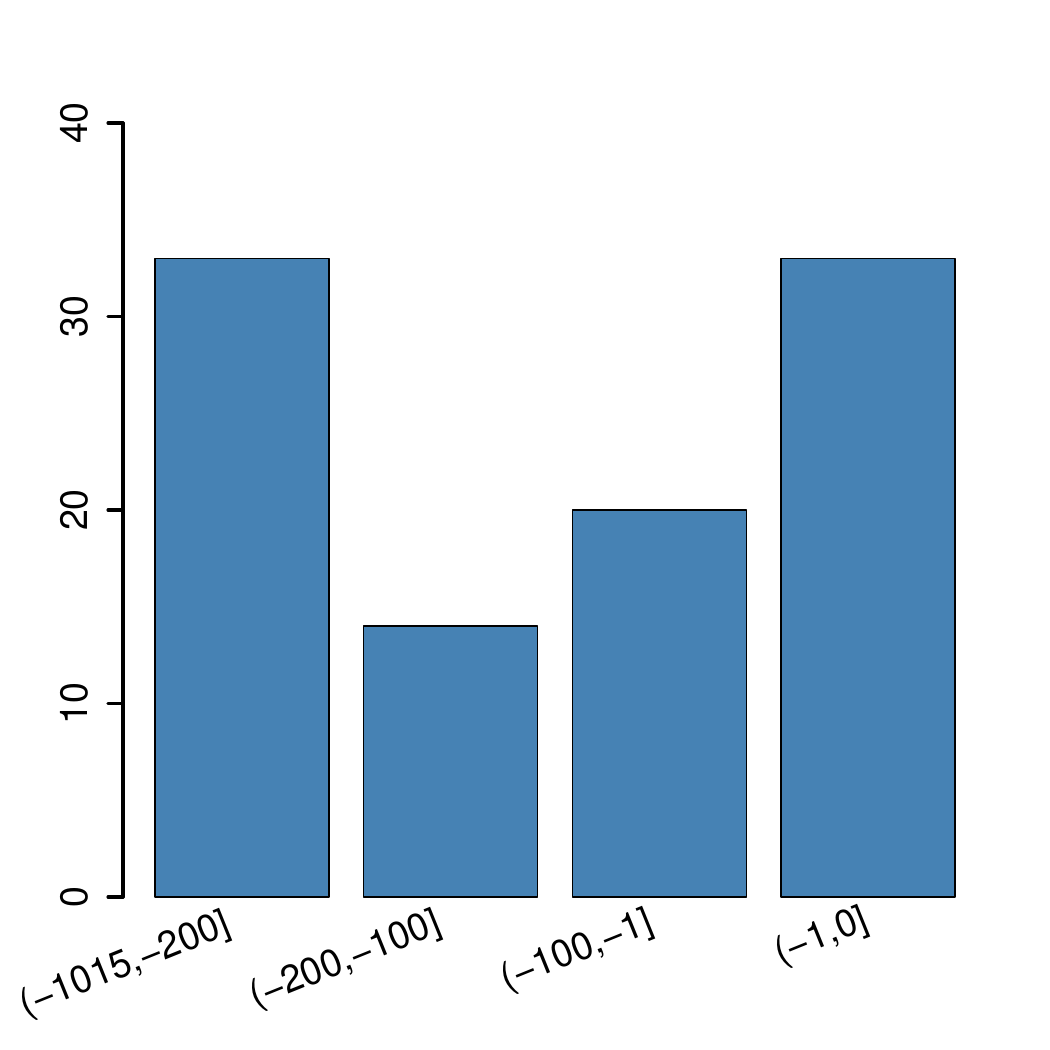}}%
	\subfloat[TWO vs. CWO]{\includegraphics[width=0.32\linewidth]{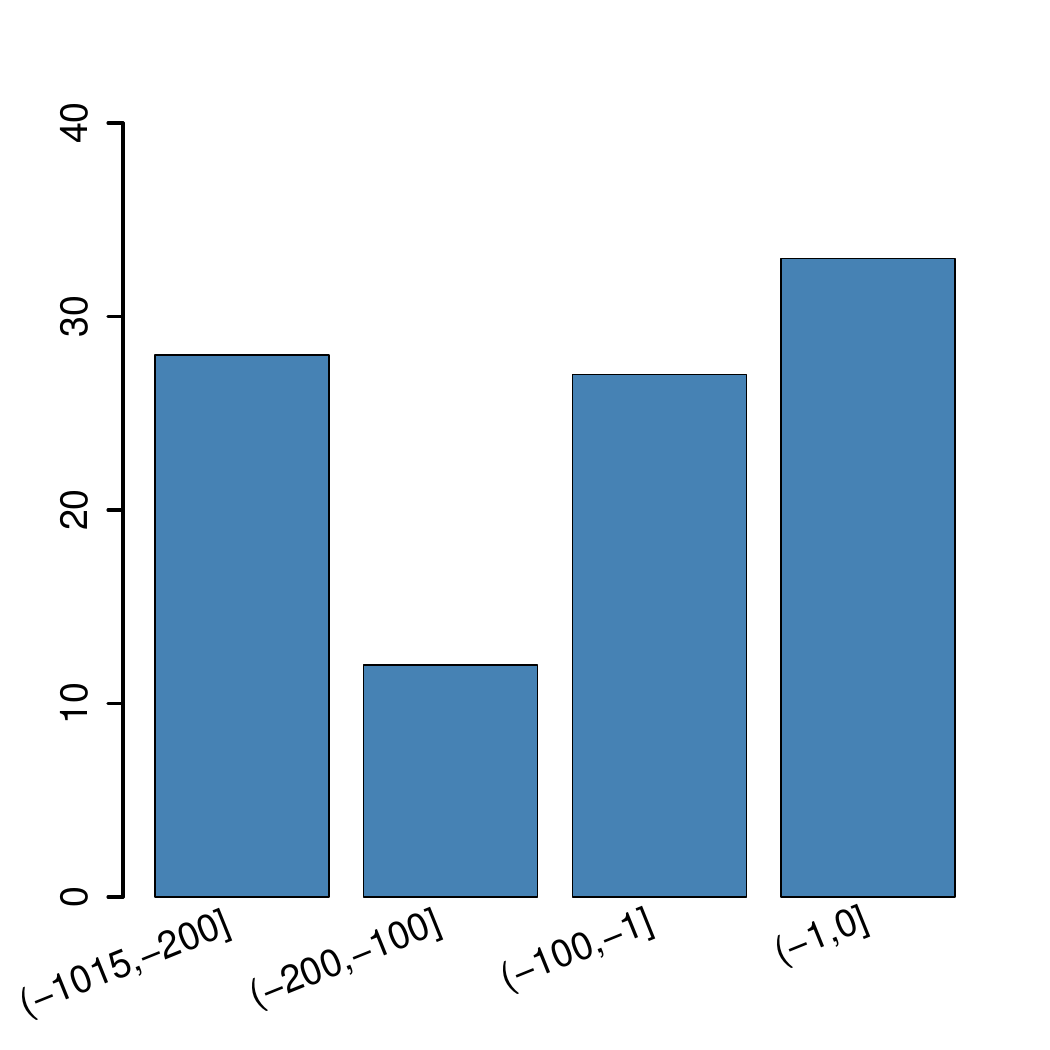}}%
	\caption{
		Histogram of reductions in PAG-list length before validation when
		comparing the original, CWO (collider-with-order), and TWO
		(triple-with-order) variants of IOD.
		The reductions are grouped into intervals $[\rm min, \rm max]$.
		Each bar shows the number of runs that had a reduction in that range.
		The reduction is computed as list size from the first method minus the
		list size from the second method, as specified in the corresponding subcaption.
	}
	\label{fig:100oracle}
\end{figure}

The simulation results align with expectations. All three approaches
produced identical final output lists containing the true PAG, confirming that the
adaptations preserve soundness and completeness. Differences arise only in the
size of the candidate list prior to validation, as shown in 
Figure~\ref{fig:100oracle}. 


Comparing the original IOD with the collider-with-order version, 63 of 100
simulations showed no change. In the remaining 37 cases, incorporating
non-ancestral relations from colliders with order consistently reduced the
number of PAGs obtained prior to validation, with a minimum reduction of 18 PAGs
and a maximum reduction of 401 PAGs.
Including all triples with order led to substantially larger reductions.
Compared with the original IOD, the triple-with-order version produced identical
lists in only 33 simulations, while the remaining 67 simulations showed reductions
ranging from a minimum of 12 PAGs to a maximum of 1,014 PAGs.
In roughly half of these cases, the reduction exceeded 200 PAGs.
This pronounced effect reflects the high frequency of identifiable non-colliders,
whether unshielded or of higher order, which can be often detected within triples.

A comparison of the triple-with-order and collider-with-order versions exhibits
a similar pattern: 33 simulations showed no differences, while the remaining
67 simulations exhibited reductions ranging from a minimum of 10 PAGs to a maximum of 1,014 PAGs.
Although the reductions are slightly smaller in a few cases,
the triple-with-order version outperforms the collider-with-order version in 67\% of simulations and
never performs worse.

In summary, simulations under faithfulness demonstrate that incorporating triples with
order produces substantially smaller lists of candidate PAGs prior to validation in most
runs. The proposed modifications improve efficiency without compromising correctness and
consistently outperforms the original IOD algorithm.

%% file: 99appendix_E.tex
\clearpage
\section{Further Simulation Results}
\label{sec:appx C}

\begin{figure}[H]
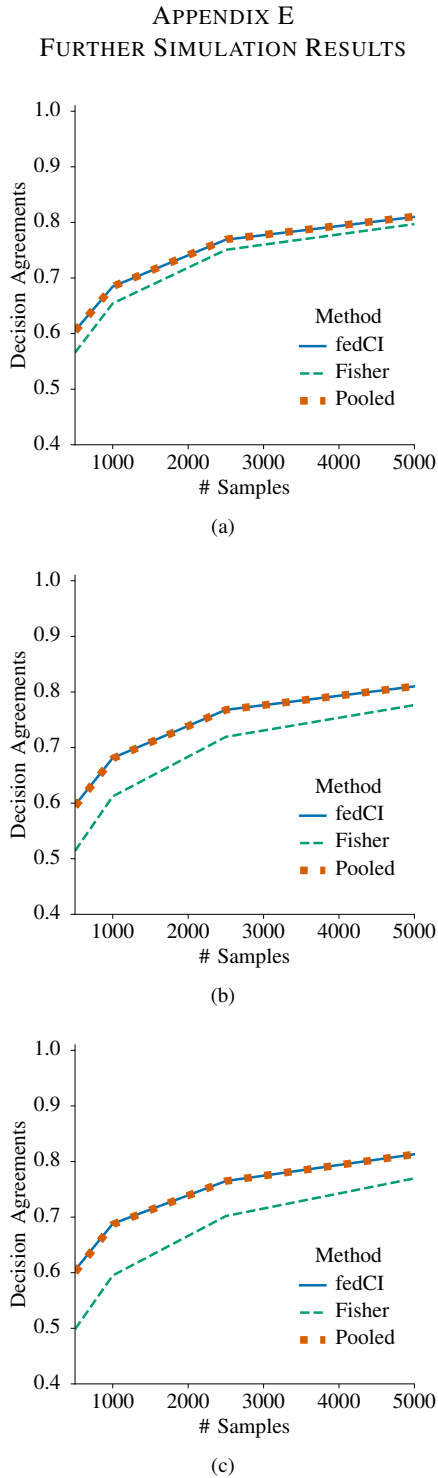

	\centering
	\subfloat[\label{fig:acc comparison 4p dep}]{\resizebox{0.65\linewidth}{!}{\includesvg[width=0.8\linewidth,inkscapelatex=true]{images/slides_acc/accuracy-dep-c4.svg}}}%
	\\
	\subfloat[\label{fig:acc comparison 8p dep}]{\resizebox{0.65\linewidth}{!}{\includesvg[width=0.8\linewidth,inkscapelatex=true]{images/slides_acc/accuracy-dep-c8.svg}}}%
	\\
	\subfloat[\label{fig:acc comparison 12p dep}]{\resizebox{0.65\linewidth}{!}{\includesvg[width=0.8\linewidth,inkscapelatex=true]{images/slides_acc/accuracy-dep-c12.svg}}}%
	\caption{
		Comparison of accuracy between fedCI Fisher's method and a pooled CI test at determining the correct relationship of conditional \textbf{dependence} between variables.
		The $y$-axis displays the decision agreements of each approach with the true m-separation or m-connectedness between pairs of variables,
		whereas the $x$-axis shows the total number of samples that was tested on.
		Figures (a), (b), and (c)
		display these results for $4$, $8$, and $12$ partitions, respectively.
		The solid line represents the results of fedCI, almost perfectly aligning with the dotted line, which represents the pooled tests, across all samples and partitionings.
		In contrast, Fisher's method, represented by a dashed line, is unable to match the performance of the other methods, especially when the number of partitions is increased.
	}
	\label{fig:acc comparison dep}
\end{figure}

\begin{figure}[H]
	\centering
	\subfloat[\label{fig:acc comparison 4p indep}]{\resizebox{0.65\linewidth}{!}{\includesvg[width=0.8\linewidth,inkscapelatex=true]{images/slides_acc/accuracy-indep-c4.svg}}}%
	\\
	\subfloat[\label{fig:acc comparison 8p indep}]{\resizebox{0.65\linewidth}{!}{\includesvg[width=0.8\linewidth,inkscapelatex=true]{images/slides_acc/accuracy-indep-c8.svg}}}%
	\\
	\subfloat[\label{fig:acc comparison 12p indep}]{\resizebox{0.65\linewidth}{!}{\includesvg[width=0.8\linewidth,inkscapelatex=true]{images/slides_acc/accuracy-indep-c12.svg}}}%
	\caption{
		Comparison of accuracy between fedCI Fisher's method and a pooled CI test at determining the correct relationship of conditional \textbf{independence} between variables.
		The $y$-axis displays the decision agreements of each approach with the true m-separation or m-connectedness between pairs of variables,
		whereas the $x$-axis shows the total number of samples that was tested on.
		Figures (a), (b), and (c)
		display these results for $4$, $8$, and $12$ partitions, respectively.
		The solid line represents the results of fedCI, almost perfectly aligning with the dotted line, which represents the pooled tests, across all samples and partitionings.
		Fisher's method, represented by a dashed line, obtains similar results, sometimes even outperforming the pooled test, likely due to its tendency to not reject the null hypothesis.
	}
	\label{fig:acc comparison indep}
\end{figure}

\begin{figure}[H]
	\centering
	\subfloat[\label{fig:logratio other sample sizes 500}]{\resizebox{\linewidth}{!}{\includesvg[width=1.1\linewidth,inkscapelatex=true]{images/logratio2pooled/by-partitions-s500.svg}}}%
	\\
	\subfloat[\label{fig:logratio other sample sizes 2500}]{\resizebox{\linewidth}{!}{\includesvg[width=1.1\linewidth,inkscapelatex=true]{images/logratio2pooled/by-partitions-s2500.svg}}}%
	\\
	\subfloat[\label{fig:logratio other sample sizes 5000}]{\resizebox{\linewidth}{!}{\includesvg[width=1.1\linewidth,inkscapelatex=true]{images/logratio2pooled/by-partitions-s5000.svg}}}%
	\caption{
		Boxplot of log-ratios of $p$-values for fedCI and Fisher's method relative to the pooled baseline on
		(a) $500$,
		(b) $2,500$,
		(c) $5,000$,
		samples, across varying partition counts.
		The fedCI results remain centered at zero with minimal variance,
		while Fisher's method shows increasing deviation and positive bias as the number of sites increases,
		especially in small sample scenarios.
		The major outliers observed for fedCI arise from numerical instability in the pooled test when multiple discrete variables are involved,
		leading to very small cell counts.
	}
	\label{fig:logratio other sample sizes}
\end{figure}


\begin{figure}[H]
	\centering
	\subfloat[\label{fig:logratio other sample sizes 500 CA}]{\resizebox{\linewidth}{!}{\includesvg[width=1.1\linewidth,inkscapelatex=true]{images/logratio2pooled/by-partitions-s500-CA.svg}}}%
	\\
	\subfloat[\label{fig:logratio other sample sizes 2500 CA}]{\resizebox{\linewidth}{!}{\includesvg[width=1.1\linewidth,inkscapelatex=true]{images/logratio2pooled/by-partitions-s2500-CA.svg}}}%
	\\
	\subfloat[\label{fig:logratio other sample sizes 5000 CA}]{\resizebox{\linewidth}{!}{\includesvg[width=1.1\linewidth,inkscapelatex=true]{images/logratio2pooled/by-partitions-s5000-CA.svg}}}%
	\caption{
		Boxplot of log-ratios of $p$-values for fedCI-CA and Fisher's method relative to the pooled baseline on
		(a) $500$,
		(b) $2,500$,
		(c) $5,000$,
		samples, across varying partition counts.
		The fedCI-CA results remain centered at zero with minimal variance,
		while Fisher's method shows increasing deviation and positive bias as the number of sites increases,
		especially in small sample scenarios.
		The major outliers observed for fedCI-CA arise from numerical instability in the pooled test when multiple discrete variables are involved,
		leading to very small cell counts.
	}
	\label{fig:logratio other sample sizes CA}
\end{figure}


\clearpage
\begin{figure}[H]
	\centering
	\resizebox{\linewidth}{!}{
		\includesvg[width=1.3\linewidth,inkscapelatex=true]{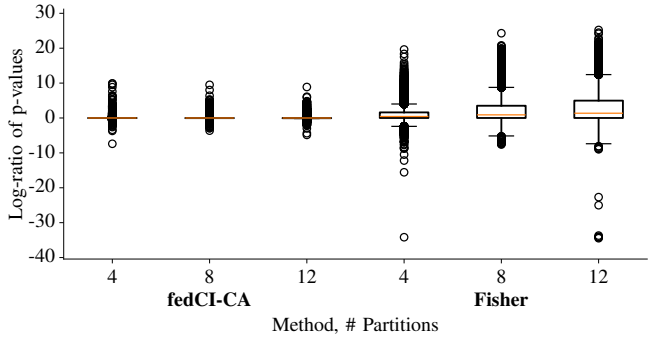}
	}
	\caption{
		Boxplot of log-ratios of $p$-values for fedCI-CA and Fisher's method relative to the pooled baseline on $1,000$ samples, across varying partition counts.
		The fedCI-CA results remain centered at zero with minimal variance, while Fisher's method shows increasing deviation and positive bias as the number of sites increases.
		This closely resembles the results of the regular fedCI.
	}
	\label{fig:logratio by sample 1000s CA}
\end{figure}


\begin{table}[H]
	\centering
	\caption{
		Average pairwise difference of best normalized SHD across simulation runs, grouped by number of samples and partitions.
		Zero indicates perfect recovery of one of the causal graphs under the restrictions of vertical partitioning.
		Similarly to fedCI, fedCI-CA clearly outperforms Fisher's method, with a consistently smaller mean differences.
	}
	\label{table:shd fedci-ca}
	\begin{tabular}{ l|r|r|r|r|r }
		\hline
		\multirow{2}{*}{Algorithm} & \multirow{2}{*}{Part.} & \multicolumn{4}{c}{Samples}                               \\
		                           &                        & $500$                       & $1,000$ & $2,500$ & $5,000$ \\
		\hline
		\hline
		\multirow{3}{*}{Pooled}    & $4$                    & $0.337$                     & $0.286$ & $0.217$ & $0.174$ \\
		                           & $8$                    & $0.348$                     & $0.289$ & $0.224$ & $0.182$ \\
		                           & $12$                   & $0.344$                     & $0.287$ & $0.213$ & $0.166$ \\
		\hline
		\multirow{3}{*}{Fisher}    & $4$                    & $0.353$                     & $0.306$ & $0.230$ & $0.190$ \\
		                           & $8$                    & $0.393$                     & $0.334$ & $0.258$ & $0.216$ \\
		                           & $12$                   & $0.403$                     & $0.344$ & $0.268$ & $0.208$ \\
		\hline
		\multirow{3}{*}{fedCI-CA}  & $4$                    & $0.335$                     & $0.287$ & $0.217$ & $0.175$ \\
		                           & $8$                    & $0.346$                     & $0.288$ & $0.224$ & $0.182$ \\
		                           & $12$                   & $0.340$                     & $0.287$ & $0.212$ & $0.166$ \\
		\hline
	\end{tabular}
\end{table}

\begin{table}[H]
	\centering
	\caption{
		Cohen's $d$ for the difference of best normalized SHDs for Pooled PAG candidates versus Fisher and fedCI-CA PAG candidates, grouped by the number of samples and partitions.
		Similarly to fedCI, fedCI-CA clearly outperforms Fisher's method, with a consistently smaller Cohen's $d$.
	}
	\label{table:shd diff fedci-ca}
	\begin{tabular}{ l|r|r|r|r|r }
		\hline

		\multirow{2}{*}{Algorithm} & \multirow{2}{*}{Part.} & \multicolumn{4}{c}{Samples}                                  \\
		                           &                        & $500$                       & $1,000$  & $2,500$  & $5,000$  \\
		\hline
		\hline
		\multirow{3}{*}{Fisher}    & $4$                    & $0.159$                     & $0.162$  & $0.103$  & $0.171$  \\
		                           & $8$                    & $0.378$                     & $0.364$  & $0.223$  & $0.302$  \\
		                           & $12$                   & $0.455$                     & $0.366$  & $0.427$  & $0.411$  \\
		\hline
		\multirow{3}{*}{fedCI}     & $4$                    & $-0.022$                    & $-0.038$ & $0.012$  & $0.081$  \\
		                           & $8$                    & $-0.083$                    & $-0.048$ & $0.000$  & $-0.039$ \\
		                           & $12$                   & $-0.117$                    & $-0.066$ & $-0.053$ & $0.039$  \\
		\hline
	\end{tabular}
\end{table}

%% file: 99appendix_F.tex
\clearpage
\onecolumn
\section{FedCI-IOD WebApp}
\label{sec:webapp appx}
\begin{figure*}[h]
	\centering
	\includesvg[width=0.8\textwidth, pretex=\tiny, inkscapelatex=true]{images/iodapp-vertical.svg}
	\caption{
		Screenshots from the fedCI-IOD web application, showcasing the individual processing steps of uploading the input data,
		connecting to a server and joining a collaboration room with other participants, as well as the final outputs: merged public PAGs and local PAGs for each client.
	}
	\label{fig:webapp 1}
\end{figure*}

%% file: bibliography.bib
@Article{richardson2002ancestral,
  author    = {Richardson, Thomas and Spirtes, Peter},
  journal   = {Ann. Statist.},
  title     = {Ancestral graph {Markov}  models},
  year      = {2002},
  number    = {4},
  pages     = {962--1030},
  volume    = {30},
  doi       = {10.1214/AOS/1031689015},
  fjournal  = {The Annals of Statistics},
  publisher = {Institute of Mathematical Statistics},
}

@book{Spirtes2001,
  author    = {P. Spirtes and C. N. Glymour and R. Scheines},
  title     = {Causation, Prediction, and Search},
  edition   = {2nd ed.},
  publisher = {MIT Press},
  address   = {Cambridge, MA, USA},
  year      = {2001},
  doi = {10.7551/mitpress/1754.001.0001}
}

@Article{Ribeiro2025,
  author  = {Ribeiro, Ad{\`e}le H and Heider, Dominik},
  journal = {arXiv preprint arXiv:2505.06542},
  title   = {{dcFCI: Robust Causal Discovery Under Latent Confounding, Unfaithfulness, and Mixed Data}},
  year    = {2025},
}

@Article{Ribeiro2020,
  author   = {Ribeiro, Adèle H. and Soler, Júlia Maria Pavan},
  journal  = {Statist. Med.},
  title    = {Learning genetic and environmental graphical models from family data},
  year     = {2020},
  number   = {18},
  pages    = {2403-2422},
  volume   = {39},
  doi      = {10.1002/sim.8545},
  eprint   = {https://onlinelibrary.wiley.com/doi/pdf/10.1002/sim.8545},
  fjournal = {Statistics in Medicine},
  url      = {https://onlinelibrary.wiley.com/doi/abs/10.1002/sim.8545},
}

@article{Chen2025,
  title={Horizontal and Vertical Federated Causal Structure Learning via Higher-order Cumulants},
  author={Chen, Wei and Gu, Wanyang and Peng, Linjun and Cai, Ruichu and Hao, Zhifeng and Zhang, Kun},
  journal={arXiv preprint arXiv:2507.06888},
  year={2025}
}

@Article{Tsagris2018,
  author    = {Tsagris, Michail and Borboudakis, Giorgos and Lagani, Vincenzo and Tsamardinos, Ioannis},
  fjournal   = {International Journal of Data Science and Analytics},
  journal   = {Int. J. Data Sci. Anal.},
  title     = {Constraint-based causal discovery with mixed data},
  year      = {2018},
  month     = {August},
  number    = {1},
  pages     = {19-30},
  volume    = {6},
  const     = {\ text},
  doi       = {10.1007/s41060-018-0097-y},
  ranking   = {rank5},
  url       = {https://doi.org/10.1007/s41060-018-0097-y http://link.springer.com/10.1007/s41060-018-0097-y},
}

@Book{Fisher1925,
  author        = {Fisher, R. A.},
  publisher     = {Edinburgh, UK: Oliver \& Boyd},
  title         = {Statistical methods for research workers},
  year          = {1925},
  comment       = {- Fishers method reference},
  doi           = {10.1093/oso/9780198522294.002.0003},
  ranking       = {rank2},
}

@Article{Zhang2008,
  author   = {Jiji Zhang},
  journal  = {Artif. Intell.},
  title    = {On the completeness of orientation rules for causal discovery in the presence of latent confounders and selection bias},
  year     = {2008},
  issn     = {0004-3702},
  number   = {16},
  pages    = {1873-1896},
  volume   = {172},
  comment  = {- explains some MAG rules and uses discriminating paths},
  doi      = {10.1016/j.artint.2008.08.001},
  url      = {https://www.sciencedirect.com/science/article/pii/S0004370208001008},
}

@Article{Chen2013,
  author   = {Chen, Colin},
  journal  = {Stat. Interface},
  title    = {Distributed iteratively reweighted least squares and applications},
  year     = {2013},
  month    = {01},
  pages    = {585-593},
  volume   = {6},
  comment  = {- Showed how to federatedly compute IRLS
- Setting was distributed computing to be faster on large datasets, not necessarily to have different clients contributing},
  doi      = {10.4310/SII.2013.v6.n4.a15},
  fjournal = {Statistics and its Interface},
}

@Misc{Wang2021,
  author        = {Lun Wang and Qi Pang and Shuai Wang and Dawn Song},
  title         = {FED-$\chi^2$: Privacy Preserving Federated Correlation Test},
  year          = {2021},
  archiveprefix = {arXiv},
  comment       = {- Fed-chi2 test
- Used in FedC2SL causal discovery

- only on},
  eprint        = {2105.14618},
  primaryclass  = {cs.CR},
  url           = {https://arxiv.org/abs/2105.14618},
  urldate = {2026-02-24},
}

@Article{Glover2004,
  author   = {Glover, Scott and Dixon, Peter},
  journal  = {Psychon. B. Rev.},
  title    = {Likelihood ratios: A simple and flexible statistic for empirical psychologists},
  year     = {2004},
  issn     = {1531-5320},
  number   = {5},
  pages    = {791--806},
  volume   = {11},
  comment  = {- Support for LRT},
  doi      = {10.3758/BF03196706},
  fjournal = {Psychonomic Bulletin & Review},
  refid    = {Glover2004},
  url      = {https://doi.org/10.3758/BF03196706},
}

@Article{McCullagh1984,
  author   = {Peter McCullagh},
  journal  = {Eur. J. Oper. Res.},
  title    = {Generalized linear models},
  year     = {1984},
  issn     = {0377-2217},
  number   = {3},
  pages    = {285-292},
  volume   = {16},
  comment  = {- Origin of GLMs},
  doi      = {10.1016/0377-2217(84)90282-0},
  fjournal = {European Journal of Operational Research},
  url      = {https://www.sciencedirect.com/science/article/pii/0377221784902820},
}

@Article{Fisher1925a,
  author   = {Fisher, R. A.},
  journal  = {Math. Proc. Cambridge Philos. Soc.},
  title    = {Theory of Statistical Estimation},
  year     = {1925},
  number   = {5},
  pages    = {700–725},
  volume   = {22},
  comment  = {- More work of Fisher on MLE},
  doi      = {10.1017/S0305004100009580},
  fjournal = {Mathematical Proceedings of the Cambridge Philosophical Society},
}

@Article{Cohen1981,
  author    = {Cohen, A. I.},
  journal   = {J. Optimiz. Theory App.},
  title     = {Stepsize analysis for descent methods},
  year      = {1981},
  issn      = {1573-2878},
  month     = feb,
  number    = {2},
  pages     = {187--205},
  volume    = {33},
  comment   = {- Explains line search and compares some methods},
  doi       = {10.1007/bf00935546},
  fjournal  = {Journal of Optimization Theory and Applications},
  publisher = {Springer Science and Business Media LLC},
}

@techreport{Gavin2016,
  author = {Gavin, Henri P.},
  title = {The {Levenberg-Marquardt} method for nonlinear least squares curve-fitting problems},
  institution = {Duke University},
  year = {2016},
  month = {September},
}

@Article{Li2025,
  author    = {Li, Yiyao and Guo, Yeting and Cao, Ligong and Wang, Haotian and Lan, Long},
  journal   = {Inform. Sciences},
  title     = {{FedPuzzle: Federated causal discovery from distributed heterogeneous variable sets}},
  year      = {2025},
  issn      = {0020-0255},
  month     = dec,
  comment   = {- FedPuzzle

- partially overlapping datasets
- assumes causal sufficiency (of union of all client vars)},
  doi       = {10.1016/j.ins.2025.123021},
  fjournal  = {Information Sciences},
  publisher = {Elsevier BV},
}

@InBook{Jabbari2017,
  author    = {Jabbari, Fattaneh and Ramsey, Joseph and Spirtes, Peter and Cooper, Gregory},
  pages     = {142--157},
  publisher = {Springer International Publishing},
  title     = {Discovery of Causal Models that Contain Latent Variables Through {Bayesian} Scoring of Independence Constraints},
  year      = {2017},
  isbn      = {9783319712468},
  fbooktitle = {Machine Learning and Knowledge Discovery in Databases},
  booktitle = {ECML PKDD},
  comment   = {- Uses PAG SHD per edge mark},
  doi       = {10.1007/978-3-319-71246-8_9},
  issn      = {1611-3349},
}

@book{Pearl2009,
  author    = {Judea Pearl},
  title     = {Causality: Models, Reasoning, and Inference},
  edition   = {2nd ed.},
  publisher = {Cambridge Univ. Press},
  address   = {Cambridge, UK},
  year      = {2009}
}

@inproceedings{Bonawitz2016,
  author    = {K. A. Bonawitz and Vladimir Ivanov and Ben Kreuter and Antonio Marcedone and H. Brendan McMahan and Sarvar Patel and Daniel Ramage and Aaron Segal and Karn Seth},
  title     = {Practical Secure Aggregation for Federated Learning on User-Held Data},
  booktitle = {NeurIPS Workshop Private Multi-Party Mach. Learn.},
  year      = {2016},
  url       = {https://arxiv.org/abs/1611.04482},
}

@Article{Gao2023,
  author  = {Erdun Gao and Junjia Chen and Li Shen and Tongliang Liu and Mingming Gong and Howard Bondell},
  fjournal = {Transactions of Machine Learning Research},
  journal = {TMLR},
  title   = {Fed{DAG}: Federated {DAG} Structure Learning},
  year    = {2023},
  comment = {citation as per their git

- score-based approach
- gradient descent/ascend based
- Additive Noise Models (ANMs)do

- Assumes causal sufficiency
- Cannot handle partially overlapping variable
- Seems to be only for continuos data (even on GitHub repo no sign of other variable types)},
  url     = {https://openreview.net/forum?id=MzWgBjZ6Le},
}

@inproceedings{Li2024,
title={Federated Causal Discovery from Heterogeneous Data},
author={Loka Li and Ignavier Ng and Gongxu Luo and Biwei Huang and Guangyi Chen and Tongliang Liu and Bin Gu and Kun Zhang},
fbooktitle={The Twelfth International Conference on Learning Representations},
booktitle={ICLR},
year={2024},
url={https://openreview.net/forum?id=m7tJxajC3G}
}

@Article{Ypma1995,
  author   = {Ypma, Tjalling J.},
  journal  = {SIAM Rev.},
  title    = {Historical Development of the {Newton–Raphson} Method},
  year     = {1995},
  number   = {4},
  pages    = {531-551},
  volume   = {37},
  comment  = {Historical analysis of origin of {Newton-Rhapson}},
  doi      = {10.1137/1037125},
  eprint   = {https://doi.org/10.1137/1037125},
  fjournal = {SIAM Review},
  ranking  = {rank2},
  url      = {https://doi.org/10.1137/1037125},
}

@Article{Yang2019,
  author    = {Yang, Qiang and Liu, Yang and Chen, Tianjian and Tong, Yongxin},
  journal   = {ACM Trans. Intell. Syst. Technol.},
  title     = {Federated Machine Learning: Concept and Applications},
  year      = {2019},
  issn      = {2157-6912},
  month     = jan,
  number    = {2},
  pages     = {1--19},
  volume    = {10},
  doi       = {10.1145/3298981},
  fjournal  = {ACM Transactions on Intelligent Systems and Technology},
  publisher = {Association for Computing Machinery (ACM)},
}

@Article{MaullinSapey2021,
  author     = {Maullin-Sapey, Thomas and Nichols, Thomas E.},
  journal    = {Stat. Comput.},
  title      = {Fisher Scoring for crossed factor linear mixed models},
  year       = {2021},
  issn       = {0960-3174},
  month      = sep,
  number     = {5},
  volume     = {31},
  address    = {USA},
  comment    = {Simplified Fisher Scoring},
  doi        = {10.1007/s11222-021-10026-6},
  fjournal   = {Statistics and Computing},
  issue_date = {Sep 2021},
  numpages   = {25},
  publisher  = {Kluwer Academic Publishers},
  url        = {https://doi.org/10.1007/s11222-021-10026-6},
}

@Article{Fisher1922,
  author    = {Fisher, R. A.},
  journal   = {Philos. Trans. R. Soc. Lond. A},
  title     = {On the mathematical foundations of theoretical statistics},
  year      = {1922},
  pages     = {309--368},
  volume    = {222},
  comment   = {- Origin of MLE},
  doi = {10.1098/rsta.1922.0009}
}

@website{lin2019simmixed,
  author = {Lin, I.},
  title = {{simMixedDAG}},
  year = {2019},
  publisher = {GitHub},
  journal = {GitHub repository},
  url= {https://github.com/IyarLin/simMixedDAG},
  commit = {11f720c104faca4834cf76b3dcb376c67b864e8a},
  urldate = {2026-02-24}
}

@inproceedings{ramsey2018tetrad,
  title={Tetrad—a toolbox for causal discovery},
  author={Ramsey, Joseph D and Zhang, Kun and Glymour, Madelyn and Romero, Ruben Sanchez and Huang, Biwei and Ebert-Uphoff, Imme and Samarasinghe, Savini and Barnes, Elizabeth A and Glymour, Clark},
  fbooktitle={8th international workshop on climate informatics},
  booktitle={CI 2018},
  pages={1--4},
  year={2018},
  organization={Boulder, CO, USA: National Center for Atmospheric Research}
}

@phdthesis{zhang2006causal,
  title={Causal inference and reasoning in causally insufficient systems},
  author={Zhang, Jiji},
  year={2006},
  school={Carnegie Mellon University}
}

@article{ali2009markov,
author = {R. Ayesha Ali and Thomas S. Richardson and Peter Spirtes},
title = {{Markov equivalence for ancestral graphs}},
volume = {37},
fjournal = {The Annals of Statistics},
journal = {Ann. Statist.},
number = {5B},
publisher = {Institute of Mathematical Statistics},
pages = {2808 -- 2837},
year = {2009},
doi = {10.1214/08-AOS626},
url = {https://doi.org/10.1214/08-AOS626}
}

@InProceedings{claassen2022greedy,
  title = 	 {Greedy equivalence search in the presence of latent confounders},
  author =       {Claassen, Tom and Bucur, Ioan G.},
  fbooktitle = 	 {Proceedings of the Thirty-Eighth Conference on Uncertainty in Artificial Intelligence},
  booktitle = 	 {Proc. 38th Conf. UAI},
  pages = 	 {443--452},
  year = 	 {2022},
  editor = 	 {Cussens, James and Zhang, Kun},
  volume = 	 {180},
  fseries = 	 {Proceedings of Machine Learning Research},
  month = 	 {01--05 Aug},
  publisher =    {PMLR},
  url = 	 {https://proceedings.mlr.press/v180/claassen22a.html}
}

@Article{Hoerl1970,
  author    = {Arthur E. Hoerl and Robert W. Kennard},
  journal   = {Technometrics},
  title     = {Ridge Regression: Biased Estimation for Nonorthogonal Problems},
  year      = {1970},
  issn      = {00401706},
  number    = {1},
  pages     = {55--67},
  volume    = {12},
  publisher = {[Taylor \& Francis, Ltd., American Statistical Association, American Society for Quality]},
  url       = {http://www.jstor.org/stable/1267351},
  doi = {10.2307/1271436}
}

@InProceedings{Tillman2011,
  author    = {Tillman, Robert and Spirtes, Peter},
  fbooktitle = {Proceedings of the Fourteenth International Conference on Artificial Intelligence and Statistics},
  booktitle = {AISTATS},
  title     = {Learning equivalence classes of acyclic models with latent and selection variables from multiple datasets with overlapping variables},
  year      = {2011},
  address   = {FL, USA},
  month     = {11--13 Apr},
  pages     = {3--15},
  fpublisher = {PMLR},
  fseries    = {Proceedings of Machine Learning Research},
  series    = {PMLR},
  volume    = {15},
  comment   = {Integration of Overlapping Datasets

- Improves on the ION Algo (Danks + Tillman)
- No local PAG computation, only local p-values},
  url       = {https://proceedings.mlr.press/v15/tillman11a.html},
}

@Article{Osborne1992,
  author    = {M. R. Osborne},
  fjournal   = {International Statistical Review / Revue Internationale de Statistique},
  journal   = {Int. Stat. Rev.},
  title     = {Fisher's Method of Scoring},
  year      = {1992},
  issn      = {03067734, 17515823},
  number    = {1},
  pages     = {99--117},
  volume    = {60},
  comment   = {Discusses Fisher scoring vs Newton

- Fisher Information is positive (semi) definite, therefore the scoring step is always downhill, always reducing the negative log-likelihood
	- Newton uses Hessian which is not necessarily positive (semi) definite, potentially causing convergence issues},
  publisher = {[Wiley, International Statistical Institute (ISI)]},
  url       = {http://www.jstor.org/stable/1403504},
  doi = {10.2307/1403504}
}

@Article{Green1984,
  author   = {Green, P. J.},
  fjournal  = {Journal of the Royal Statistical Society: Series B (Methodological)},
  journal  = {J. R. Stat. Soc. Ser. B},
  title    = {Iteratively Reweighted Least Squares for Maximum Likelihood Estimation, and Some Robust and Resistant Alternatives},
  year     = {1984},
  issn     = {0035-9246},
  month    = {12},
  number   = {2},
  pages    = {149-170},
  volume   = {46},
  doi      = {10.1111/j.2517-6161.1984.tb01288.x},
  eprint   = {https://academic.oup.com/jrsssb/article-pdf/46/2/149/49172867/jrsssb\_46\_2\_149.pdf},
  url      = {https://doi.org/10.1111/j.2517-6161.1984.tb01288.x},
}

@InBook{Wang2023,
  author    = {Wang, Zhaoyu and Ma, Pingchuan and Wang, Shuai},
  pages     = {351–367},
  publisher = {Springer Nature Switzerland},
  title     = {Towards Practical Federated Causal Structure Learning},
  year      = {2023},
  isbn      = {9783031434150},
  fbooktitle = {Machine Learning and Knowledge Discovery in Databases: Research Track},
  booktitle = {ECML PKDD},
  comment   = {- FedC2SL

- Chi2 test on discrete variables by pooling contigency table},
  doi       = {10.1007/978-3-031-43415-0_21},
  issn      = {1611-3349},
  url       = {http://dx.doi.org/10.1007/978-3-031-43415-0_21},
}

@inproceedings{Zhang2007,
author = {Zhang, Jiji},
title = {A Characterization of {Markov} Equivalence Classes for Directed Acyclic Graphs with Latent Variables},
year = {2007},
isbn = {0974903930},
publisher = {AUAI Press},
address = {Arlington, Virginia, USA},
fbooktitle = {Proceedings of the Twenty-Third Conference on Uncertainty in Artificial Intelligence},
booktitle = {Proc. 23rd Conf. UAI},
pages = {450–457},
numpages = {8},
location = {Vancouver, BC, Canada},
url           = {https://arxiv.org/abs/1206.5282},
}

@InProceedings{Hannun2021,
  author    = {Hannun, Awni and Guo, Chuan and van der Maaten, Laurens},
  fbooktitle = {Proceedings of the Thirty-Seventh Conference on Uncertainty in Artificial Intelligence},
  booktitle = {Proc. 37th Conf. UAI},
  title     = {Measuring data leakage in machine-learning models with Fisher information},
  year      = {2021},
  month     = {27--30 Jul},
  pages     = {760--770},
  publisher = {PMLR},
  fseries    = {Proceedings of Machine Learning Research},
  volume    = {161},
  comment   = {- Quantifies information loss of methods that use Fisher information: Fisher Information Loss (FIL)},
  url       = {https://proceedings.mlr.press/v161/hannun21a.html},
}

@Article{Dwork2014,
  author    = {Dwork, Cynthia and Roth, Aaron},
  journal   = {Foundations and Trends® in Theoretical Computer Science},
  title     = {The Algorithmic Foundations of Differential Privacy},
  year      = {2014},
  issn      = {1551-3068},
  month     = aug,
  number    = {3–4},
  pages     = {211--487},
  volume    = {9},
  doi       = {10.1561/0400000042},
  publisher = {Emerald},
}

@Article{Tajabadi2024,
  author    = {Tajabadi, Mohammad and Martin, Roman and Heider, Dominik},
  journal   = {Comput. Struct. Biotechnol. J.},
  title     = {Privacy-preserving decentralized learning methods for biomedical applications},
  year      = {2024},
  issn      = {2001-0370},
  month     = dec,
  pages     = {3281--3287},
  volume    = {23},
  doi       = {10.1016/j.csbj.2024.08.024},
  publisher = {American Association for the Advancement of Science (AAAS)},
}

@Article{Hauschild2022,
  author    = {Hauschild, Anne-Christin and Lemanczyk, Marta and Matschinske, Julian and Frisch, Tobias and Zolotareva, Olga and Holzinger, Andreas and Baumbach, Jan and Heider, Dominik},
  journal   = {Bioinformatics},
  title     = {Federated Random Forests can improve local performance of predictive models for various healthcare applications},
  year      = {2022},
  issn      = {1367-4811},
  month     = feb,
  number    = {8},
  pages     = {2278--2286},
  volume    = {38},
  doi       = {10.1093/bioinformatics/btac065},
  editor    = {Wren, Jonathan},
  publisher = {Oxford University Press (OUP)},
}
